\documentclass{article} 
\usepackage[preprint]{colm2026_conference}

\usepackage{microtype}
\usepackage{hyperref}
\usepackage{url}
\usepackage{booktabs}

\usepackage{float}
\usepackage{placeins}
\usepackage{lineno}
\usepackage{hyperref}
\usepackage{url}
\usepackage{booktabs,multirow}
\usepackage[table]{xcolor} 
\usepackage{xspace}
\usepackage{graphicx}
\usepackage{subcaption}
\usepackage{paralist}
\usepackage{booktabs}  
\usepackage{multirow}   
\usepackage{tabularx}
\usepackage[pass]{geometry}
\usepackage{hyperref}
\usepackage{xcolor}
\usepackage{ragged2e}
\usepackage{tcolorbox}
\usepackage{wrapfig}
\usepackage{setspace}
\usepackage{multicol} 
\usepackage{enumitem}
\usepackage{soul}
\usepackage{soulutf8}
\usepackage{adjustbox}
\usepackage{ulem}
\usepackage[bottom]{footmisc}
\usepackage{pifont}
\usepackage{nameref}
\definecolor{darkblue}{rgb}{0, 0, 0.5}
\hypersetup{colorlinks=true, citecolor=darkblue, linkcolor=darkblue, urlcolor=darkblue}

\tcbset{
  promptbox/.style={
    enhanced,
    breakable,  
    colback=boxbg,
    colframe=boxborder,
    colbacktitle=titleblue,
    coltitle=white,
    fonttitle=\bfseries,
    boxrule=0.8pt,
    arc=4pt,
    left=8pt,right=8pt,top=8pt,bottom=8pt
  }
}

\title{When Cases Get Rare: A Retrieval Benchmark for\\Off-Guideline Clinical Question Answering}

\author{Doeun Lee$^{1}$, Muge Zhang$^{1}$, Yi Yu$^{1}$, \\
\textbf{Ashish Manne}$^{2}$\textbf{, Stephen Koesters}$^{2}$\textbf{, Frank Wen}$^{3}$\textbf{, Brady Buchanan}$^{2}$\textbf{,} \\
\textbf{Lynda Villagomez}$^{2}$\textbf{, Oluwatoba Moninuola}$^{2}$\textbf{, James Lim}$^{2}$\textbf{, Kathryn Tobin}$^{2}$\textbf{,} \\
\textbf{Andrew Srisuwananukorn}$^{2}$\textbf{, Ping Zhang}$^{1}$\textbf{, Sachin Kumar}$^{1}$ \\
\\
$^{1}$The Ohio State University \\
$^{2}$The Ohio State University Wexner Medical Center \\
$^{3}$ University of Chicago Medical Center \\
}

\ifcolmsubmission
\linenumbers
\fi

\newif\ifshowcomments
\showcommentsfalse

\newcommand{\sk}[1]{\ifshowcomments \textcolor{blue}{[#1 -SK]}  \fi}

\newcommand{\dl}[1]{\ifshowcomments\textcolor{orange}{[#1 -DL]}\fi}

\newcommand{\Sref}[1]{\S\ref{#1}}

\newcommand{\benchmark}[0]{\textsc{OGCaReBench}\xspace}

\tcbset{
  txtbox/.style={
    colback=white,
    colframe=black,
    arc=0pt,
    boxrule=0.6pt,
    left=8pt,right=8pt,top=6pt,bottom=6pt,
    enlarge left by=0pt,
  }
}
\begin{document}
\maketitle

\begin{abstract}
Across medical specialties, clinical practice is anchored in evidence-based guidelines that codify best studied diagnostic and treatment pathways. These pathways 
 routinely fall short for the long tail of real-world care not covered by guidelines. Most medical large language models (LLMs), however, are trained to encode common, guideline-focused medical knowledge in their parameters. Current evaluations test models primarily on recalling and reasoning with this memorized content, often in multiple-choice settings. Given the fundamental importance of evidence-based reasoning in medicine, it is neither feasible nor reliable to depend on  memorization in practice. To address this gap, we introduce \benchmark, a free-form retrieval-focused benchmark aimed at evaluating LLMs at answering clinical questions that require going beyond typical guidelines. Extracted from published medical case reports and validated by medical experts, \benchmark contains long-form clinical questions requiring free-text answers, providing a systematic framework for assessing open-ended medical reasoning in rare, case-based scenarios. Our experiments reveal that even the best-performing baseline (GPT-5.2) correctly answers only 56\% of our benchmark with specialized models only reaching 42\%. Augmenting models with retrieved medical articles improves this performance to up to 82\% (using GPT-5.2) highlighting the importance of evidence-grounding for real-world medical reasoning tasks. This work thus establishes a foundation for benchmarking and advancing both general-purpose and medical LLMs to produce reliable answers in challenging clinical contexts.

\end{abstract}

\section{Introduction}

Large language models are actively being explored in healthcare settings for many use cases with potential to transform clinical decision-making and ultimately enhance patient outcomes \citep{yan2024largelanguagemodelbenchmarks, abrar2024empiricalevaluationlargelanguage,shool2025systematic}. Realizing this potential requires evaluations that reflect the diversity and complexity of real clinical scenarios. 
Most current benchmarks, however, test models' recall of medical knowledge through exam-style questions  \citep{Ben_Abacha_2019medquad, Krithara2023bioasqqa}, 
typically in multiple-choice settings \citep{jin2019pubmedqa, medqa, pal2022medmcqa, hendrycks2021MMLU, zuo2025medxpertqa}. While free-form question-answering datasets exist, they are largely patient-oriented and not designed for clinician-facing decision support \citep{hosseini2024benchmarklongformmedicalquestion, nguyen-etal-2023-medredqa, Singhal2023HealthSearchQA, zhu-etal-2020-mashqa}. 
Furthermore, evidence grounding is especially crucial in medicine, where clinical guidance evolves rapidly, authoritative references are essential for trust, and patient care often involves rare conditions and atypical presentations. Memorization alone is thus insufficient; models must be able to integrate and synthesize knowledge from external sources to support real-world clinical decision-making.

We aim to benchmark LLMs in settings that reflect how physicians approach complex clinical problems.
Such a benchmark must satisfy three key properties: (1) it should be grounded in real patient cases reflecting the variability and nuance of clinical practice, (2) it should adopt a free-form question-answering format to capture the open-ended reasoning physicians require, as opposed to multiple choice questions, and (3) it should be non-trivial, demanding expert-level domain knowledge, mirroring the complexity of real-world decision-making. 
Guided by these principles, we focus on simulating scenarios in which physicians must consult external resources to determine appropriate clinical decisions for patients whose cases fall outside standard guidelines or involve rare, off-guideline presentations. 

\begin{figure}
    \centering
    \includegraphics[width=1.0\linewidth]{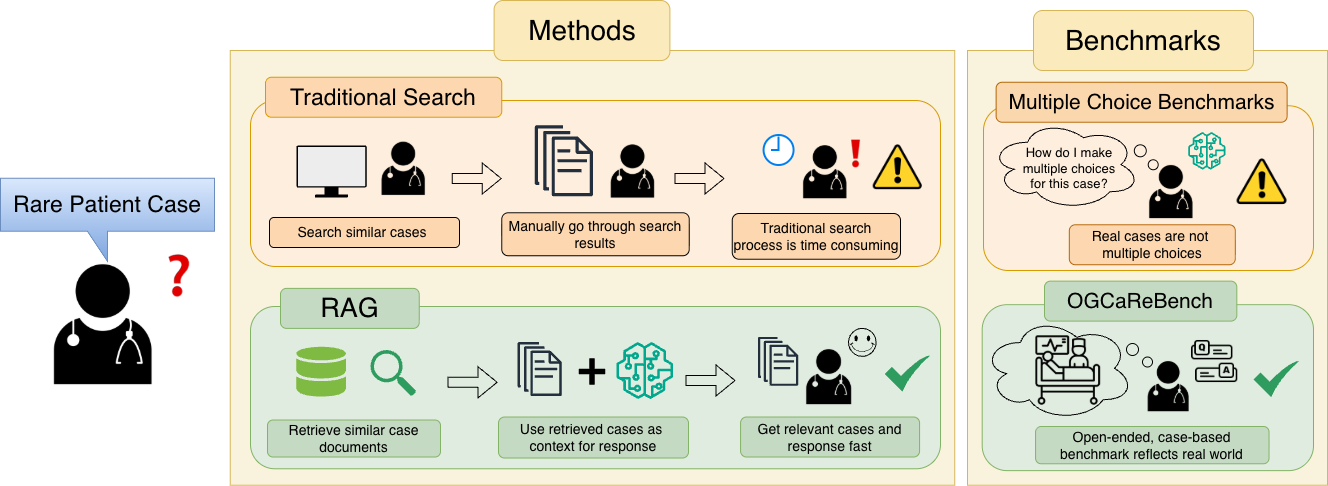}
        \caption{Physicians facing rare clinical cases that fall outside standard medical guidelines often search for similar prior cases to guide diagnosis or treatment. Traditional search requires manually reviewing many results; retrieval augmented LLMs can use relevant case documents as context to provide faster, case-grounded assistance. Existing medical LLM benchmarks are largely multiple-choice and do not reflect this open-ended clinical workflow. \benchmark evaluates LLMs on rare, case-based clinical questions to fill this gap.}
    
    \label{fig:fig1}
\end{figure}
For this purpose, we use published medical case reports. Case reports document novel, rare, or unprecedented clinical occurrences such as unusual case presentations, atypical diagnostic mechanisms or non-standard treatments. Physicians often consult them when typical guideline references, such as UpToDate \citep{UpToDate2025WK} or standard specialty guidelines, are insufficient to manage complex or unusual cases. 
For each case report, we apply semi-automatic methods (\Sref{sec:benchmark}) to extract a question and answer pair centered around the significant contribution of the report---which could be a novel diagnosis, novel treatment, or a test associated with a rare occurrence of a disease. 
We refer to this medical benchmark of \textbf{O}ff-\textbf{G}uideline \textbf{Ca}se \textbf{Re}ports as \benchmark. Our dataset contains 
cases across 10  medical specialties (see \autoref{tab:case_reports}). 
All questions and answers are validated by experienced physicians to ensure accuracy and fidelity to real-world clinical reasoning. 

Our evaluation of several state-of-the-art general-purpose and medical domain-specific models reveals that LLMs struggle to provide expected responses to rare cases. 
These results highlight the limitations of relying on parameteric memory of the models alone when handling rare cases, underscoring the necessity of retrieval augmentation in complex medical scenarios. Therefore, we expand our horizon to evaluating performances under retrieval, which 
is known to enhance the performance of medical question answering \citep{RagInMed}. 
We create a retrieval corpus of 53,617 case reports covering 12 medical specialties, drawn from publicly available reports. We find providing retrieved documents in the context of the question significantly increases model performance
. While proprietary models generally outperform open-source models, they are not perfect. 
Even with oracle document provided, they continue to make significant number of errors, indicating limitations not only from retrieval but also in correctly integrating evidence into clinical questions. 

%
In summary, we make the following contributions: 
\begin{itemize}[leftmargin=*]
    \item We introduce \benchmark, an expert-validated benchmark derived from published medical case reports, to evaluate language models on realistic rare clinical scenarios.
    \item We empirically demonstrate the shortcomings of both medical and general-purpose models in open-ended rare-case reasoning, underscoring the limitations of their standalone use for supporting physicians in real clinical settings.
    \item We show that retrieval augmentation enhances performance in expert-level tasks, emphasizing its necessity in building robust systems in the medical domain. 
\end{itemize}
\section{Related Work} 
\textbf{Models and Datasets Focused on Medicine}
Medical question-answering (QA) models have significantly evolved \citep{shool2025systematic, yan2024largelanguagemodelbenchmarks}. A large portion focuses on and is mostly tested on multiple-choice question answering \citep{han2025medalpacaopensourcecollection, wu2023pmcllamabuildingopensourcelanguage,
singhal2023expertlevelmedicalquestionanswering, bolton2024biomedlm27bparameterlanguage}, often using exam-style benchmarks for evaluation \citep{shool2025systematic, Krithara2023bioasqqa, jin2019pubmedqa, medqa, pal2022medmcqa, hendrycks2021MMLU, zuo2025medxpertqa}. Models and datasets with free-form answers are often patient-oriented \citep{li2023chatdoctormedicalchatmodel, hosseini2024benchmarklongformmedicalquestion,nguyen-etal-2023-medredqa,Singhal2023HealthSearchQA,zhu-etal-2020-mashqa} or based on general clinical knowledge
\citep{GarciaFerrero2024MedMT5AO,bolton2024biomedlm27bparameterlanguage,Krithara2023bioasqqa} rather than case-conscious reasoning. Nevertheless, there is increasing interest in more complex free form QA in medical domain recently \citep{healthbench_openai_2025, hicks2026healthbenchprofessionalevaluatinglarge}.
Especially, there have been studies focusing on case-based models and datasets \citep{chen2026rarearena}, many of them focusing on reasoning \citep{xu2025reversephysicianairelationshipfullprocess, nori2025sequentialdiagnosislanguagemodels}.
\citet{zhao2026agentic} presents an agent system with reasoning traces for rare disease diagnosis. Most related to our work are 
\citet{Qiu2025QuantifyingTR} and \citet{wu2025medcasereasoningevaluatinglearningdiagnostic}, which also focus on using case reports to construct a benchmark for final diagnosis, clinical reasoning, and treatments. However, majority of these evaluations are done with very limited expert validation which limits their trustworthiness.\sk{i modified the previous sentence a bit, if i misrepresented the facts, please feel free to revert.}\dl{I removed "synthetic" because they are from case reports too} Our work, while similarly constructed, introduces modifications (see \autoref{subsec:Step 3}) to the case scenario to ensure the cases presented are unprecedented. The full dataset is annotated to confirm the modifications. We provide a comparison table with other benchmarks involving expert annotation on \autoref{tab:comparison}.
Additionally, we broaden the focus and convey the novelty of the case report, whether it may be diagnosis, treatment, or clinical examinations that are presented in a novel way.

{\renewcommand{\arraystretch}{1.3}
\begin{table}
\centering
\small
\setlength{\tabcolsep}{0.3pt}
\begin{tabular}{lcccc}
\toprule
\textbf{Benchmark} & \textbf{Size (Annotated)} & \textbf{Format} & \textbf{Case Based?} & \textbf{Level} \\
\midrule
\textbf{PubMedQA-Test Set} \citep{jin2019pubmedqa} & 500 (500) & Binary & \ding{55} & Expert \\
 
\textbf{HealthSearchQA} \citep{Singhal2023HealthSearchQA} & 3173 (100) & Question Only & \ding{55} & Consumer \\

\textbf{MedCaseReasoning} \citep{wu2025medcasereasoningevaluatinglearningdiagnostic} & 14489 (100) & Stepwise Diagnosis & \ding{51} & Expert \\
 
\textbf{SDBench} \citep{nori2025sequentialdiagnosislanguagemodels} & 304 (56) & Stepwise Diagnosis & \ding{51} & Expert \\
 
\textbf{RareArena} \citep{chen2026rarearena} & $\sim$50000 (100) & Diagnosis & \ding{51} & Expert \\
\textbf{HealthBench} \citep{healthbench_openai_2025} & 5000 (5000) & Free Form & \ding{55} & Both \\
\textbf{HealthBenchPro} \citep{hicks2026healthbenchprofessionalevaluatinglarge} & 525 (525) & Free Form & \ding{55} & Expert \\
\textbf{MedBrowseComp} \citep{chen2025medbrowsecompbenchmarkingmedicaldeep} & $\sim$1100 ($\sim$1100) & Short Answer & \ding{55} & Expert \\ 
\midrule
\textbf{\benchmark} (Ours) & 639 (639) & Free Form & \ding{51} & Expert \\
\bottomrule
\end{tabular}
\caption{Comparison table of popular benchmark datasets in the medical domain that involve expert annotation. \benchmark provides the largest free form, cased-based dataset with full expert annotation among comparable medical benchmarks. \dl{should i state names and description of each benchmark on related work section?}}
\label{tab:comparison}
\end{table}}
\textbf{Retrieval augmentation in expert domains}
Retrieval augmentation generation (RAG) is known to enhance the performance in knowledge-intensive tasks \citep{lewis2021retrievalaugmentedgenerationknowledgeintensivenlp}, providing a promising foundation for domain-specific reasoning \citep{Lee2025ChainofRankEL}. Using RAG in areas requiring domain expertise mitigates the limitation of memorization by integrating curated professional context as shown by examples from legal domains \citep{Zheng_2025, hou2024clercdatasetlegalcase}. With medicine, prior studies have shown that incorporating RAG enhances the performance in various medical QA, ranging from multiple choice to case-based reasoning \citep{xiong2024benchmarkingretrievalaugmentedgenerationmedicine,
Dong2025TalkBY, Ke2025, chen2025medbrowsecompbenchmarkingmedicaldeep, Jeong2024ImprovingMR}. There is also an increasing interest in agentic models that use retrieval internally such as OpenEvidence \citep{openevidence2024} and Deep-DxSearch \citep{zheng2026endtoendagenticragtraining}. However, use of RAG targeting various rare-case scenarios and case-based retrieval corpus still remains a gap, and we address this by evaluating rare-case questions using RAG.\sk{would this need to updated too to incorporate the agentic models which use retrieval internally? we can also cite "open evidence" platform here.}\dl{done}

\section{\benchmark{}: A benchmark of off-guideline medical cases}\label{sec:benchmark}

Medical case reports document novel or rare clinical occurrences. They are typically published to document and highlight unusual conditions, atypical disease courses, unexpected complications, new diagnosis mechanisms, or unique treatment strategies. Case reports appear in specialty journals such as, Journal of Clinical Case Reports, British Medical Journal (BMJ) Case Reports, general medical journals like New England Journal of Medicine (NEJM), and online repositories. 
To better understand how case reports are used in practice by physicians, we first conducted informal interviews with 10 physicians from different US based institutions with specialties ranging from emergency medicine, rheumatology, internal medicine, infectious diseases, oncology, and surgery. We surmised that while not all practitioners rely on case reports---fields like infectious diseases or emergency medicine rarely need to consult them---specialties such as surgery, internal medicine, and oncology often turn to case reports. Physicians reported that when encountering cases that fall outside standard clinical guidelines,\footnote{Clinical guidelines set by major societies like American College of Cardiology (ACC/AHA), the American College of Rheumatology (ACR) and the National Comprehensive Cancer Network (NCCN), and others codify large bodies of evidence and are regularly updated by broad expert panels.} they rely on case reports and series alongside consultation with colleagues or specialty networks to identify relevant precedents and guide their clinical decision-making. This is supported by studies showing that only 55\% to 57\% of guideline-recommended treatments are implemented in routine practice \citep{mcglynn2003quality,runciman2012caretrack}.

\begin{figure}
    \centering
    \includegraphics[width=1\linewidth]{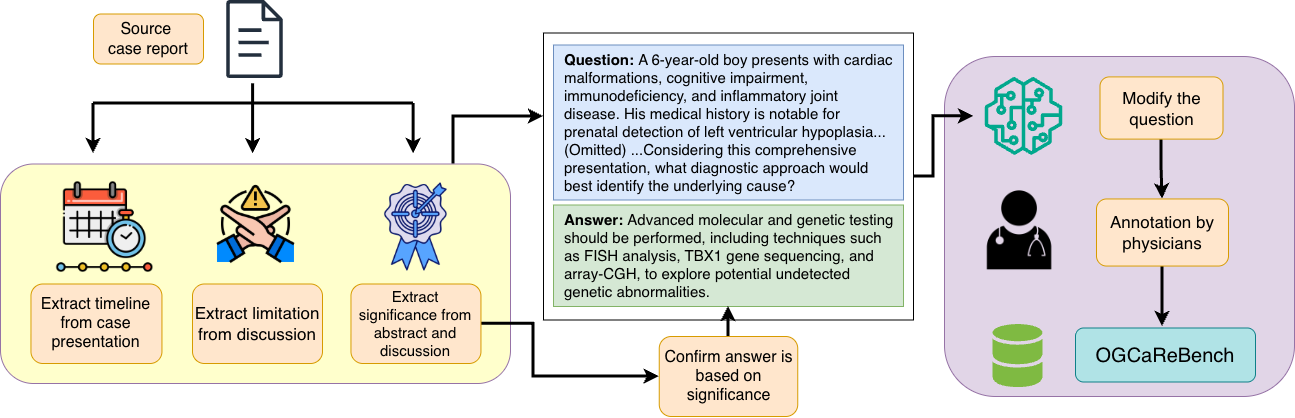}
    \caption{\benchmark{} creation pipeline.
    }
    \label{fig:dataset-construction}
\end{figure}

To construct a dataset that emphasizes such rare, patient-specific cases, we synthesize \benchmark from these reports. Starting from all open-access case reports available on PubMed Central \citep{PMCOA}, we filter for cases with novel content and persistent rarity, then extract question-answer pairs using LLMs. To simulate realistic clinical scenarios beyond the scope of the original reports, we apply controlled modifications to these questions, ensuring they are distinct from the documented cases. Finally, all modified questions undergo physician annotation to validate both accuracy and clinical relevance. We outline the benchmark construction in \autoref{fig:dataset-construction} and detail it below. An example of a case report along with the created question-answer pair is provided in \autoref{fig:case_report_excerpt}.



\subsection{Dataset Creation}\label{sec:datast_creation}
\paragraph{Step 1: Collecting and filtering case reports}
We start by collecting all open-access medical articles through PubMed Central (PMC) \citep{PMCOA}. 
We then leverage the open-access journal list provided by PMC and filter for case reports based on the venues in which these journals are published, focusing on those known to regularly feature case reports  
(see \autoref{tab:venues}). 
This results in a total of 53,617 reports. 
We use this corpus as the data store for retrieval-augmented evaluation.
To construct a dataset centered on rare cases, 
we filter this collection to remove reports which meet any of the following criteria: 
\begin{inparaenum}[(1)]
    \item the case report was published in or before 2022.
    \item more than three articles cite the case report.
    \item the case report is cited by more than one non-case report article.
\end{inparaenum}
The rationale for these filters is as follows: recently published case reports retain novelty for a few years; case reports cited fewer than a conservative three times are assumed to represent persisting rarity; and case reports cited multiple times by non-case report articles are excluded, as these follow-up studies are indicators that the case has been further explored and potentially resolved into a standard guideline. This setup may introduce false negatives, as low citation counts may reflect low visibility rather than true persisting rarity. Such cases are later filtered during expert annotation.
Based on these criteria, we obtain 28,219 reports. 

\paragraph{Step 2: Extracting raw question-answer pairs}\label{subsec:Step 2}
Among the filtered case reports, we randomly select a subset of 1,500 to constitute the dataset. Using GPT-5.2 (\texttt{2025-12-11}), we extract three key elements from each report: (1) the case presentation as a timeline providing a sequence of procedures in patient care, (2) limitations that hinder the application of standard medical procedures, motivating the use of off-guideline approaches, and (3) the most significant contribution of the report, defined as the rationale for its publication. Reports whose significance reflects solely the application of standard procedures to a rare disease, rather than the introduction of a novel intervention, are discarded as the resulting questions and answers will be trivial. The question is then formulated by presenting all procedures preceding the decision point that reflect the significance, asking for the next appropriate step given the limitations. 
For example, the most significant contribution of a report might be in developing a new diagnostic test for a condition where all standard diagnostics are inconclusive. In this case, the question will include the patient's history up until the point where the new test was performed, along with unavailable diagnostics based on the limitations, and ask what the next step should be. This procedure simulates a scenario where a physician encounters a similar case and has run out of standard guideline-recommended options. The corresponding answer for this question would be the subsequent significant step which in this example is the new diagnostic test. Finally, we use the LLM to verify that the answer was directly connected to the identified significance, thereby confirming the integrity of the question-answer generation process. 

\paragraph{Step 3: Adding distractors to generated questions}\label{subsec:Step 3}
Questions generated in Step 2 are further modified to increase their realism. As our goal is to simulate a situation where physicians consult case reports as guidance for treating their own patients, it is essential that the questions represent unforeseen scenarios and are presented differently from the original reports from which they were derived. To achieve this, we introduce controlled modifications---referred to as distractions---using Claude 4 Opus. These modifications include altering patient demographics (e.g., age and ethnicity), substituting medical terminology with semantically equivalent expressions, integrating comorbidities that do not affect the original condition, and other related adjustments (see \autoref{fig:prompt_distractions} for the full prompt). Importantly, while the questions are modified, the answers are preserved; distractions are applied only to the extent that the clinical plausibility of the case remains intact and the correct answer remains unaffected. To validate this process, three internal medicine physicians are presented with subsets of the original question, modified question, and corresponding answer from an early version of the dataset. 
Their evaluation confirms the medical coherence of the modifications. 
Such distrations mirror the challenges physicians face in real-world settings where unrelated comorbidities may exist.

\paragraph{Step 4: Dataset verification by experts}
We assess the medical validity of the question-answer pairs using annotations from eight physicians in internal medicine or pediatrics. The experts are presented with the modified questions and asked to evaluate them based on the following criteria: 
(1) the question and answer should be medically aligned, and (2) the question should require domain-specific medical expertise rather than general medical knowledge held by the public. 
We ask them to rate the pairs on a likert scale of 1 to 5---1 indicating that question-answer pair is not realistic under any circumstances and 5 indicating that the question is realistic and the answer correctly addresses the question.
Detailed instructions we provide the physicians are in \autoref{fig:instructions}. 
Only question-answer pairs rated 4 or 5 are retained, yielding 639 instances in the final dataset. We do a bootstrap analysis to confirm that our benchmark shows stable results (see Appendix \ref{app:bootstrap}). 

\subsection{Data Statistics}
We summarize the dataset statistics split across medical specialties in \autoref{tab:case_reports}. We use the original corpus for all 53617 case reports extracted in Step 1 as our retrieval store. 
The case reports collected are represented by 12 disciplines.  
Both the full set and \benchmark{} are heavily inclined toward Internal Medicine and Surgical Studies. This imbalance is expected and reflects the specialties where this benchmark is most applicable.
For internal medicine, this is due to its overlap with other specialties and it also encompassing a variety of sub-disciplines such as hepato-biliary-pancreatic and vascular medicine. 
\begin{wraptable}[19]{r}{0.49\textwidth}
\centering
\small
\begin{tabular}{lrr}
\toprule
\textbf{Specialty} & \begin{tabular}{@{}c@{}}\textbf{All} \\ \textbf{Reports}\end{tabular} & \begin{tabular}{@{}c@{}}\textbf{OGCaRe-} \\ \textbf{Bench}\end{tabular} \\
\midrule
Basic Sciences & 4571 & 25 \\
Dentistry & 1678 & 14 \\
External Health & 3394 & 27 \\
Intensive Care & 2699 & 34 \\
Internal Medicine & 15606 & 284 \\
Neurology & 1872 & 17 \\
OBGYN & 1654 & 11 \\
Oncology & 1078 & 12 \\
Orthopedics & 2646 & 32 \\
Pediatrics & 1642 & 14 \\
Surgical Studies & 13216 & 169 \\
General Medicine & 1396 & - \\
Others & 2165 & - \\
\midrule
Total & 53617 & 639 \\
\bottomrule
\end{tabular}
\caption{Distribution of case reports across specialties}
\label{tab:case_reports}
\end{wraptable}
For surgical studies, each case is unique and hence more case reports are written about them. 
Each question in \benchmark{} consists of 1-2 paragraphs, and answers are often 1-2 sentences (length distribution is reported in \autoref{tab:tokennum}).


\subsection{Evaluation Metric}
To evaluate the performance of a model using \benchmark, we feed the question to the model with an instruction to generate a free form natural language answer. To evaluate the alignment between the gold answers extracted from the case reports and the responses generated by the models, we use an LLM-as-a-judge to assess equivalence (specifically, we use GPT-5.2). In our early experiments, we find that model responses have varying formats and lengths, ranging from brief phrases to long paragraphs that include background and rationale. To focus on the main clinical content, we prompt the judge with a few-shot example to output ``equivalent'' or ``mismatch'' (see \autoref{fig:prompt_eval} for full prompt). Model response is judged as equivalent if the primary clinical procedure recommended matches the procedure specified in the gold answer.  Conversely, a response is considered a mismatch if the contents do not overlap or the gold answer appears in the output but not prioritized as the main procedure. Similarly, broad or vague recommendations are labeled as mismatches when the gold answer requires a specific clinical procedure, as our benchmark emphasizes detailed, case-based clinical reasoning. Our primary metric is a simple percentage of answers in the benchmark predicted correctly by the model. To validate this LLM-based evaluation, we randomly select 100 baseline results evenly spread across all models and setup to be validated by internal medicine physicians. We task them to label whether the GPT-5.2's evaluation of matching model-generated answers and gold answers reflects true clinical judgment, yielding an agreement of 93\%. 
\section{Evaluation Setup}
We consider two setups, (1) a baseline setup in which a model is expected to rely on its own parametric knowledge,  and (2) a setup where we first perform retrieval on our datastore to find the most relevant case reports and provide the retrieved documents to the model's context to generate the answer. For the retrieval augmented generation (RAG) setup, we first validate the performance of different retrieval models and use the top-performing ones for final evaluation.

\subsection{Baseline Evaluation}
We benchmark six state of the art general-purpose and three medical domain models as baselines. The full list is on Appendix \ref{app:model_full}.
 We prompt the models to answer the question with one best answer (see \autoref{fig:prompt_a}). Restrictions such as not outputting thoughts and a word limit are added to medical QA models to avoid having unusually lengthy output (see \autoref{fig:prompt_b}). We explored additional medical domain models but excluded from final evaluation due to their misalignment with our task (see Appendix \ref{app:model_dropped}).

\subsection{Retrieval Augmented Evaluation}

\paragraph{Evaluating Retrieval Methods}

To identify the most effective retrieval models for our downstream generation task, we evaluate a comprehensive set of 15 models encompassing sparse, general purpose, and biomedical models. We also experiment with two-stage retrieval process. Following the initial retrieval, we rerank the top 100 candidates using the PubMed-pretrained MedCPT-cross-encoder \citep{Jin_2023}, which has demonstrated state-of-the-art performance on biomedical information retrieval tasks.
To assess performance of retrieval, we report results using Recall@k, MRR, and nDCG with respect to the ground-truth case report (from which the question and answer are derived), which together capture different aspects of retrieval effectiveness. Instruction used for BMRetriever is in \autoref{fig:bm_prompt_query}. We explore various chunking methods to optimize retrieval performance, and ultimately use a maximum length of 512 tokens and a stride of 128. Full chunking method details are in Appendix \ref{app:chunking}\sk{remember to fix}.

\paragraph{Retrieval Augmented Generation}
We select the best-performing retrievers (see \autoref{fig:retrieval_graph}) from each of the three categories---sparse (BM25), general purpose (BGE), and biomedical (BMRetriever)---to incorporate into retrieval augmented evaluation. Each of the nine LLM used in baseline experiments is integrated into the pipeline. Deep research models are included with the retrieval corpus as research datastore. 
We evaluate the model performance using the top 1, 3, and 5 retrieved case reports as context, as well as an oracle setting in which the ground-truth source case report of the question is input. For OpenbioLLM and Llama 3-Med42, which have a small context window of 8K tokens, we are only able to test with a maximum of 3 retrieved reports, given that the average length of our case reports is 2,730 tokens (see \autoref{tab:tokennum}).


\section{Results and Analyses} \label{sec:result}

\begin{wraptable}[16]{r}{0.40\textwidth}
\centering
\setlength{\tabcolsep}{3pt}

\small
\begin{tabular}{l r} 
\toprule
\textbf{Model} & \textbf{Accuracy} \\
\midrule
GPT-5.2 & \textbf{56.0} \\
GPT-o4-mini & 51.8 \\
Llama 3.3 70B Instruct & 44.9 \\
Claude 4.5 Sonnet & 51.2 \\
Thinking Claude 4 Sonnet & 48.2 \\
Gemini 2.5 Pro & 46.6 \\
MedGemma-27b-text-it & 37.1 \\
Llama 3-Med42-70B & 42.1 \\
OpenBioLLM-Llama 3-70B & 37.6 \\

\bottomrule
\end{tabular}

\caption{Overall baseline performance. Subfield-level results are provided in Appendix \ref{app:baseline_subfield}}
\label{tab:baseline}
\end{wraptable}
\paragraph{Baselines without retrieval struggle} 
\autoref{tab:baseline} shows the baseline performance of the base models evaluated with \benchmark{} without retrieval augmentation. 
Surprisingly, general-purpose models overall outperform medical specialized ones. 
For example, MedGemma, the latest offering from Google, performs on the lower end at 37.1\%. 
These results show that both state-of-the-art general-purpose models and models for medical tasks struggle when presented with complex rare medical questions.
Subpar baseline performance suggests that memorization from pretraining alone is insufficient for handling such cases. 
The performance of GPT-5.2 and GPT-o4-mini suggests that reasoning offers some advantage in handling rare, case-based scenarios, while Thinking Claude 4 Sonnet doesn't follow this trend. We also speculate that OpenAI's models' performance could be due to recent efforts in improving health-related information communication \citep{OpenAI2025GPT52, healthbench_openai_2025} which may include training with domain specific data. 

\begin{figure}
    \centering
    \includegraphics[width=1\linewidth]{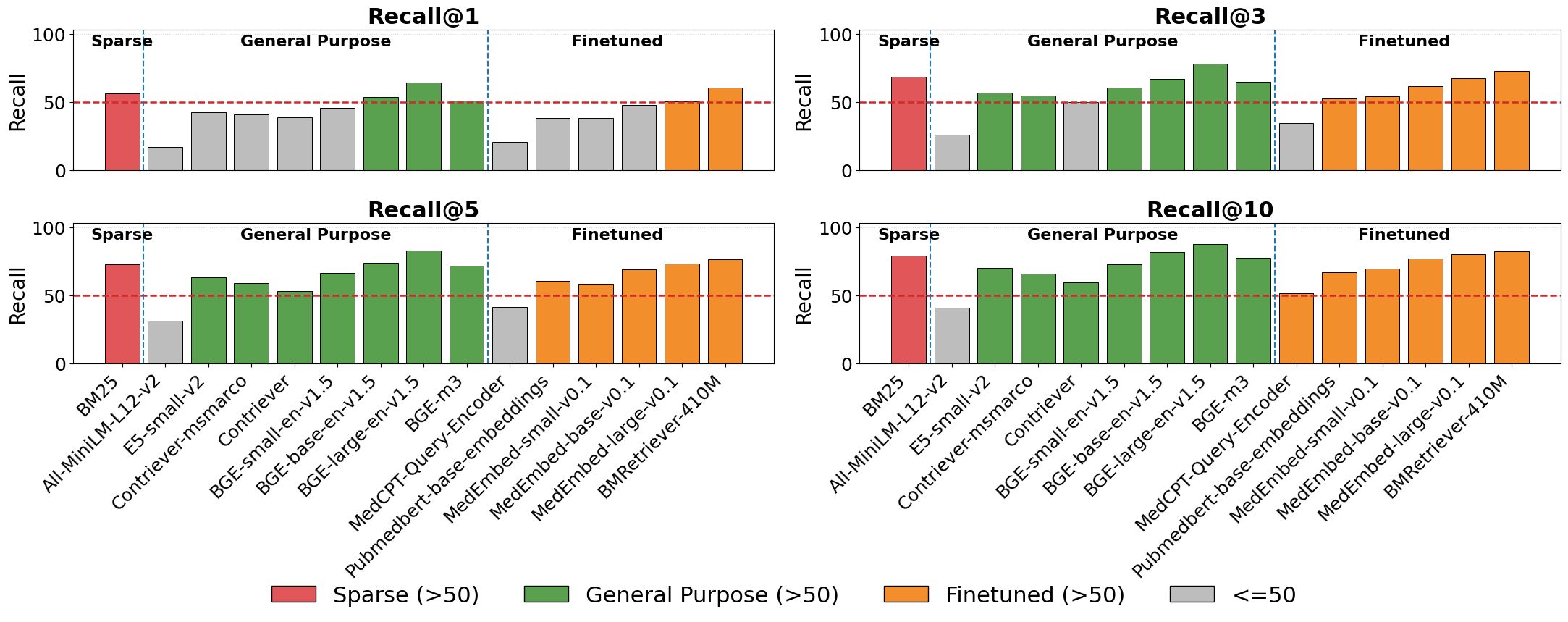}
    \caption{Retrieval result for all retriever models tested based on Recall@k measured by percentage. Sparse (red), general purpose (green), and medical finetuned (orange) models were tested. Models with accuracy less than 50\% are colored grey. Full retrieval result is in Appendix \ref{app:full_ret}
    }
    \label{fig:retrieval_graph}
\end{figure}

\paragraph{Retrieving the right document is hard for complex medical queries}
\autoref{fig:retrieval_graph} presents the retrieval performance of state-of-the-art retrievers. The results demonstrate that \benchmark{} is a challenging retrieval benchmark, with most models achieving Recall@1 below 50\%. For RAG systems, this indicates that the retrieved context is likely to miss crucial information more than half of the time, thereby reducing the quality and accuracy of generated answers. Although Recall@k approaches 100\% at very high values of $k$ (100--1,000), providing such a large number of documents as context to LLMs is impractical. Full retrieval result with other metrics are in Appendix \ref{app:full_ret} and results with simple truncation are reported in Appendix~\ref{app:retrieval truncation}\sk{check}.

\begin{wraptable}[9]{r}{0.40\textwidth}
\centering
\setlength{\tabcolsep}{3pt}

\small
\begin{tabular}{l r} 
\toprule
\textbf{Model} & \textbf{Accuracy} \\
\midrule
DR Tulu & 19.9 \\
GPT-o4-mini-deep-research & \textbf{53.5} \\

\bottomrule
\end{tabular}

\caption{Deep research models' performance.\sk{this table is not referenced or discussed. It can be discussed in RAG paragraph.}}
\label{tab:dr}
\end{wraptable}
\paragraph{Retrieval augmentation improves results for large context models, but gap remains in others.}
GPT-5.2 and GPT-o4-mini are two models with the highest performance across all three context sizes with performance reaching up to $\sim$82\%. With a context window of 5 reports (retrieved with BGE-large), GPT-5.2 even 
surpasses the oracle report performance of specialized medical models. 
Among medical models, MedGemma exhibits the most notable improvement; Llama 3-Med42 performs comparably. \autoref{tab:dr} shows that deep research models exhibit substantially lower performance compared to conventional RAG\sk{not sure what this sentence means? }. \dl{fixed, but can we remove this sentence since I mention dr tulu right after?} DR Tulu performs the worst at 19.9\%, introducing a gap in general purpose agent's ability in domain specific tasks. 
Overall, we find three important aspects that affect model performance: 
\begin{figure}
    \centering
    \includegraphics[width=1\linewidth]{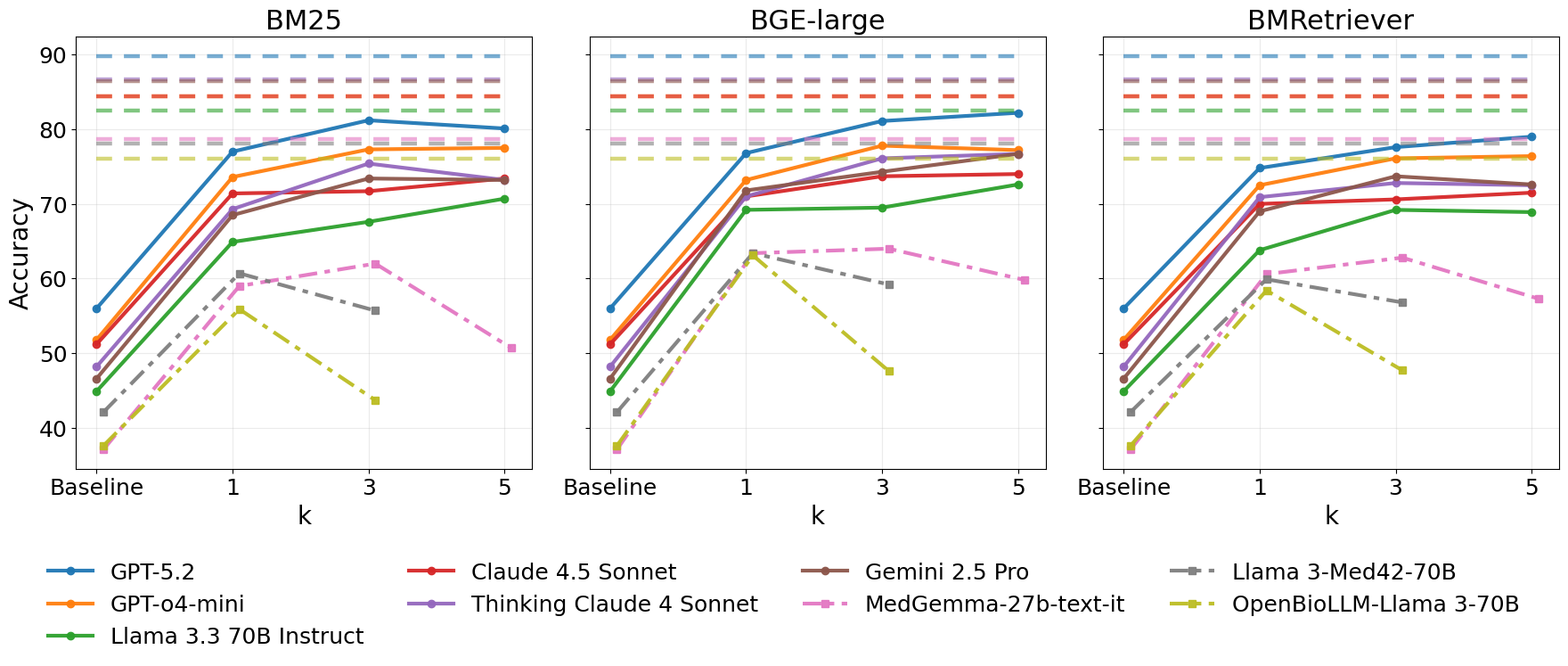}
    \caption{Performance of RAG in \% accuracy with different retrieval methods and context lengths. Oracle performances are represented with dashed lines at the top of each graphs. Claude 4.5 Sonnet and GPT-o4-mini has similar oracle accuracy, as well as Thinking Claude 4 Sonnet and Gemini 2.5 Pro. Full RAG result is in Appendix \ref{app:full_rag}.
    }
    \label{fig:rag_img}
\end{figure}
\begin{inparaenum}[(1)]
\item \textbf{Retrieval performance is a critical factor influencing RAG performance.} Comparing the language models across retrievers, using BGE-large yields the highest accuracy in most models regardless of performance level, reflecting its highest retrieval quality.
\item \textbf{Context window size, as well as the number of documents in context affects the performance.} Models with limited context capacity (Llama 3-Med42, OpenBioLLM) of 8K tokens exhibit the lowest gains when using three case reports as context. In contrast, MedGemma shows significant improvements with RAG due to its large context window, even though it has the lowest baseline performance.
\item Finally, \textbf{model's reasoning ability effects RAG performance.} 
GPT-5.2 and GPT-o4-mini achieve the best result across all retrievers and context lengths, followed by Thinking Claude 4 Sonnet and Gemini 2.5 Pro most of the time. This underscores the importance of reasoning in consulting external sources. 
\end{inparaenum}
Overall, RAG improves the performance of the models and thus proves essential for rare-case reasoning, transforming subpar baseline performance into clinically significant results.

\begin{table*}[h]
\centering
\small

\setlength{\tabcolsep}{5pt}
\begin{tabularx}{\linewidth}{>{\raggedright\arraybackslash}p{3.0cm} >{\raggedright\arraybackslash}X >{\raggedright\arraybackslash}p{3.2cm}}
\toprule
\textbf{Failure mode} &
\textbf{Diagnostic question} &
\textbf{Failure axis} \\
\midrule

Document grounding &
Did the model fail to extract, preserve, or utilize the article-supported answer? &
Source grounding \\

\addlinespace

Objective misalignment &
Did the model optimize for a different task than the question requires, although it may understand the oracle article? &
Task objective \\

\addlinespace

Granularity mismatch &
Did the model answer the right question, but at the wrong level of specificity? &
Answer specificity \\

\addlinespace

Context/stage misbinding &
Did the model select an action from the wrong stage of the workflow? &
Workflow stage \\

\addlinespace

Constraint/qualifier erosion &
Did the model omit a required qualifier, condition, or component? &
Answer completeness \\

\bottomrule
\end{tabularx}
\caption{Taxonomy of failure modes used to classify model errors on \benchmark (oracle setting).
Each case is assigned one primary and at most one secondary failure mode by a judge LLM
(GPT-5.1).\sk{this table and the heatmap could go in the appendix, perhaps. We can only define the errors you are going into details of in the main paper.}}
\label{tab:failure-modes}
\end{table*}

\paragraph{Perfect retrieval does not close the oracle gap}
The gains from RAG show that access to relevant case reports is important for rare case reasoning. However, the oracle results also show that retrieval is not the only bottleneck. Even when the oracle case is provided directly as reference, model still makes a significant number of errors. To better understand these particular failures, we define a number of failure modes as shown in \autoref{tab:failure-modes} and use an LLM judge (with manual spot checks) to classify each mismatch into one primary mode and one optional secondary mode. 

We observe two major patterns as shown in Appendix \ref{app:failure_mode}.
First, across general purpose frontier models, the dominant failure mode is model performing a different task than what the question requires, although it may
understand the oracle article (i.e., \textbf{objective misalignment}). This accounts for more than 40\% of errors for Claude 4.5 Sonnet, Gemini 2.5 Pro, GPT-5.2, and Thinking Claude 4 Sonnet.  
Rather than producing the oracle-specified action, models bundle multiple interventions, append subsequent therapeutic or procedural steps beyond what the article designates, or add unprompted concurrent interventions. 
Second, models failing to extract, preserve, or utilize
the article-supported answer (i.e., \textbf{document grounding}) and model omitting a required qualifier,
condition, or component (i.e., \textbf{constraint/qualifier erosion}) are substantial across all models. Document grounding accounts for 35.3\% of primary errors in MedGemma and 28.6\% in Llama 3-Med42, while OpenBioLLM shows particularly high rates of constraint/qualifier erosion 
and granularity mismatch 
. 
The distribution also varies by model family. Frontier models show a higher tendency to generalize beyond the article's canonical answers, while specialized models show weaker extraction of article specific decisions and poorer preservation of nuanced answer constraints. OpenBioLLM additionally shows larger share of granularity mismatch, indicating difficulty in matching required specificity of the benchmark target.
These findings show that 
improving retrieval alone is unlikely to eliminate all errors without improving grounded decision fidelity. 
\section{Conclusion}
Our work argues that reliable medical LLMs must move beyond memorization and towards benchmarks that reflect real-world clinical reasoning. \benchmark{} highlights rare, case-based scenarios where current models fall short\sk{incomplete sentence?}. Retrieval augmentation fills this gap by curating the cases to focus on, exhibited by significantly enhanced performance. Together, \benchmark{} shows retrieval as a crucial component for building clinically reliable LLMs and establishes a new benchmark for supporting physicians when faced with uncommon clinical cases. Retrieval performance, context window, number of documents used as context, and the reasoning ability all play essential roles when it comes to RAG. We hope this benchmark expands the field of open-ended rare-case reasoning in the medical domain and thereby supports physicians. We discuss limitations and avenues for future work in \autoref{app:limitations}.

\bibliography{colm2026_conference}
\bibliographystyle{colm2026_conference}
\newpage
\appendix
\section*{Appendix}
\section{LLM Usage Disclosure}

We used LLM during the creation of \benchmark{} and during evaluation. \benchmark{} was created by extracting information such as case presentation, significant point, and limitations from the source case report using GPT-5.2. Modifying the query was done by Claude Opus 4. The model output of \benchmark{} was evaluated using LLM-as-a-judge, with GPT-5.2. 

\section{Reproducibility Statement}
We provide the \benchmark{} dataset as supplementary material in csv and json format. ``Title" is the title of the source case report that the question-answer pair was derived from, ``pmc\_id" is PMC ID of the source case report, and ``Classification" indicates its specialty. The prompts for dataset construction process in \Sref{sec:datast_creation} are in \autoref{app:prompts}, including significance and timeline extraction, question-answer pair creation, controlled modification, model prompts for evaluation, and evaluating answer and model response equivalence. The full dataset and code will be publicly released upon publication. 

\section{Limitations}
\label{app:limitations}
\benchmark{} has a few limitations. First, the case reports are published and updated regularly, and it is likely that our filtered subset of 28,219 reports, as well as the cases included in \benchmark{}, will no longer satisfy the filter criteria as medical knowledge advances. This might make the rare cases part of the guideline reducing the utility of the benchmark in a few years. Therefore, future efforts may require updating the datasets or developing a dynamic version of \benchmark{} to mitigate data depreciation. This issue of benchmark saturation is however not unique to medical domain and has been explored extensively in the literature. It has additional challenges in medicine as expert annotation is required. Second, case reports vary in quality; some recommend the use of a specific product on behalf of a company. While we exclude such instances in \benchmark{}, they remain in the retrieval corpus. This may lead to the recommendation of certain products during RAG. Third, certain questions may have more than one clinically appropriate next step, and it is feasible that those are not captured by our answer. Although physician validation reduces this risk, multiple answers may still exist in some instances, given the complex nature of clinical practice. Finally, there are other venues to handle rare cases, such as platforms only accessible by physicians. They are the edge cases that are not covered by using case reports as our sources, and in practice RAG will not be able to solve them, as it is not included in any publicly available retrieval corpus. 

\section{Disclaimer} OpenBioLLM is included as one of our baselines as it has been included in multiple prior studies \mbox{\citep{shoham2024medconceptsqaopensourcemedical, dorfner2024biomedicallargelanguagesmodels}}.  However, the model is released without an accompanying paper, data description, or detailed methodological description. Therefore, its performance should be interpreted with consideration and we include it for completeness and comparability with existing benchmarks. 

\newpage
\section{Models} 
\subsection{Full List of Models Used}\label{app:model_full}
For general-purpose models, we evaluate: GPT-5.2\sk{which 5?} \dl{done}\citep{OpenAI2025GPT52}, GPT-o4-mini \citep{OpenAI2025o3o4mini},  Llama 3.3 (70B Instruct) \citep{llama33modelcard}, Claude 4.5 Sonnet \citep{Anthropic2025Claude45}, Thinking Claude 4 Sonnet \citep{Anthropic2025Claude4}, and Gemini 2.5 Pro \citep{gemini25_thinking_2025}. 
For models specializing in medical question answering, we evaluate: OpenbioLLM-Llama 3 (70B) \citep{OpenBioLLMs}, MedGemma (27B-text) \citep{sellergren2025medgemmatechnicalreport}, and Llama 3-Med42 (70B) \citep{med42v2}. 
These six general-purpose models and three medical models are used for both baseline and RAG. In addition to this, we evaluate deep research models using retrieval corpus as datastore. DR Tulu\citep{shao2025drtulureinforcementlearning} and GPT-o4-mini-deep-research \citep{o4mini_deep_research_api_2025} were tested, with web search enabled for GPT-o4-mini-deep-research. We include these models to attest to the importance of our retrieval store in solving this benchmark.
\sk{do we not do any web-search here? in any case, this should technically go after retrieval because it uses a datastore.}. \dl{added mentioning of web search}

\subsection{Retrieval Models}\label{app:models_ret_fulll}
For the sparse baseline, we employ BM25 \citep{bm25}, a model known for its strong performance across various benchmarks, including BEIR \citep{thakur2021beirheterogenousbenchmarkzeroshot}. Our general-purpose models include All-MiniLM-L12-v2\footnote{\url{https://huggingface.co/sentence-transformers/all-MiniLM-L12-v2}.}, E5-small-v2 \citep{wang2024textembeddingsweaklysupervisedcontrastive}, Contriever and Contriever-MSMARCO \citep{izacard2022unsuperviseddenseinformationretrieval}, and the BGE family  \citep{xiao2024cpackpackedresourcesgeneral}, which integrates dense, sparse, and multi-vector strategies. For the biomedical domain, we assess MedCPT \citep{Jin_2023}, PubMedBERT \citep{Gu_2021}, MedEmbed series \citep{balachandran2024medembed}, and BMRetriever\citep{xu2024bmretrievertuninglargelanguage}, a medically pre-trained and fine-tuned instruction-following model.\sk{again, i think we can move the whole list to appendix and just mention their properties, sparse, dense etc.}

\subsection{Additional Models Explored}\label{app:model_dropped}
 We considered Meditron (70B) \citep{chen2023meditron70b} and Clinical Camel (70B) \citep{toma2023clinicalcamelopenexpertlevel}, but we decided unsuitable since their primary focus is multiple-choice question answering and therefore generated responses by always outputting answer choices when given questions from \benchmark{} (even though no choices were provided in the question). \sk{this last sentence thing about meditron etc can be moved to the appendix with a short reference. I also think, this whole list of models can be moved to the appendix since you already have the list in table 3. You can just say we do closed-source and open-source and also medical models, whole list in table 3.} \dl{done}
\section{Ablation Study}
\subsection{Baseline in Subfields}\label{app:baseline_subfield}
\begin{table}[h]
\resizebox{\textwidth}{!}{
\begin{tabular}{lccccccccccc}
\toprule
\multicolumn{1}{c}{\textbf{Model}} & \multicolumn{1}{c}{\textbf{Basic Sci.}} & \multicolumn{1}{c}{\textbf{Dent.}} & \multicolumn{1}{c}{\textbf{Ext. Health}} & \multicolumn{1}{c}{\textbf{Intensive}} & \multicolumn{1}{c}{\textbf{Int. Med.}} & \multicolumn{1}{c}{\textbf{Neuro.}} & \multicolumn{1}{c}{\textbf{OBGYN}} & \multicolumn{1}{c}{\textbf{Onc.}} & \multicolumn{1}{c}{\textbf{Ortho.}} & \multicolumn{1}{c}{\textbf{Peds.}} & \multicolumn{1}{c}{\textbf{Surgical}} \\
\midrule
GPT-5.2 & 72.0 & 57.1 & 63.0 & 61.8 & 56.7 & 23.5 & 72.7 & 58.3 & 50.0 & 57.1 & 53.3 \\
GPT-o4-mini & 64.0 & 57.1 & 66.7 & 64.7 & 53.5 & 23.5 & 36.4 & 33.3 & 31.3 & 50.0 & 50.9 \\
Llama 3.3 70B Instruct & 56.0 & 57.1 & 37.0 & 58.8 & 47.5 & 17.7 & 18.2 & 33.3 & 40.6 & 35.7 & 43.2 \\
Claude 4.5 Sonnet & 52.0 & 28.6 & 63.0 & 52.9 & 54.2 & 35.3 & 36.4 & 33.3 & 46.9 & 64.3 & 49.1 \\
Thinking Claude 4 Sonnet & 48.0 & 35.7 & 48.2 & 64.7 & 52.1 & 23.5 & 36.4 & 33.3 & 28.1 & 57.1 & 46.8 \\
Gemini 2.5 Pro & 68.0 & 42.9 & 40.7 & 55.9 & 50.4 & 17.7 & 36.4 & 41.7 & 34.4 & 42.9 & 43.2 \\
MedGemma-27b-text-it & 36.0 & 35.7 & 48.2 & 50.0 & 38.0 & 11.8 & 27.3 & 25.0 & 28.1 & 28.6 & 37.9 \\
Llama 3-Med42-70B & 56.0 & 42.9 & 37.0 & 52.9 & 44.0 & 17.7 & 18.2 & 41.7 & 34.4 & 42.9 & 40.8 \\
OpenBioLLM-Llama 3-70B & 48.0 & 28.6 & 25.9 & 47.1 & 40.5 & 17.7 & 27.3 & 33.3 & 25.0 & 28.6 & 37.9 \\
\bottomrule
\end{tabular}
    }
\caption{Baseline performance across specialties. }
\label{tab:baseline_discipline}
\end{table}
\autoref{tab:baseline_discipline} shows baseline results for all specialties. Most models show better performance than the baseline for basic science and Intensive Care, but struggle to perform well in Neurology and Orthopedics. Between the two representative disciplines, Internal Medicine and Surgical Studies, all models exhibit better accuracy for internal medicine. 
\clearpage
\subsection{Full Retrieval Result}\label{app:full_ret}

\begin{table}[h]
\centering

\setlength{\tabcolsep}{1.3pt}   
\small
\resizebox{\textwidth}{!}{    
\begin{tabular}{clcrrrrrrrr}
\toprule
\multicolumn{1}{c}{\textbf{Type}} & \multicolumn{1}{c}{\textbf{Model}} & \multicolumn{1}{c}{\textbf{Params}} & \multicolumn{1}{c}{\textbf{R@1}} & \multicolumn{1}{c}{\textbf{R@3}} & \multicolumn{1}{c}{\textbf{R@5}} & \multicolumn{1}{c}{\textbf{R@10}} & \multicolumn{1}{c}{\textbf{R@100}} & \multicolumn{1}{c}{\textbf{R@1000}} & \multicolumn{1}{c}{\textbf{MRR@5}} & \multicolumn{1}{c}{\textbf{nDCG@10}} \\ \midrule
{Sparse} & BM25 & N/A & 56.0 & 68.7 & 72.6 & 79.3 & 91.1 & 95.9 & 62.6 & 67.3 \\
\midrule
\multirow{8}{*}{General}& All-MiniLM-L12-v2 & 33.4M & 16.7 & 25.7 & 31.5 & 40.8 & 70.6 & 88.4 & 21.9 & 27.3 \\
 & E5-small-v2 & 33.4M & 42.6 & 56.7 & 63.2 & 70.3 & 87.6 & 96.6 & 50.2 & 55.8 \\
 & Contriever-msmarco & 110M & 40.7 & 54.5 & 58.8 & 66.0 & 85.6 & 96.2 & 47.8 & 52.9 \\
 & Contriever & 110M & 38.5 & 49.8 & 53.2 & 59.2 & 82.5 & 95.6 & 44.4 & 48.5 \\
 & BGE-small-en-v1.5 & 33.4M & 45.4 & 60.7 & 66.2 & 72.8 & 90.8 & 98.9 & 53.6 & 58.9 \\
 & BGE-base-en-v1.5 & 109M & 53.8 & 67.1 & 73.9 & 82.0 & 92.8 & 99.4 & 61.3 & 67.1 \\
 & BGE-large-en-v1.5 & 335M & 64.3 & 78.1 & 82.9 & 87.5 & 96.4 & 99.4 & 71.6 & 75.9 \\
 & BGE-m3 & 560M & 51.0 & 64.8 & 71.7 & 77.6 & 92.3 & 98.0 & 58.6 & 63.8 \\
 \midrule
\multirow{6}{*}{Finetuned} & MedCPT-Query-Encoder & 109M & 20.8 & 34.7 & 41.2 & 51.3 & 83.1 & 97.3 & 28.3 & 34.8 \\
 & Pubmedbert-base-embeddings & 109M & 38.2 & 52.6 & 60.3 & 66.8 & 87.2 & 97.5 & 46.4 & 51.9 \\
 & MedEmbed-small-v0.1 & 33.4M & 37.9 & 54.0 & 58.4 & 69.5 & 89.4 & 98.1 & 46.1 & 52.8 \\
 & MedEmbed-base-v0.1 & 109M & 47.9 & 61.5 & 69.2 & 77.0 & 94.1 & 99.5 & 55.6 & 61.5 \\
 & MedEmbed-large-v0.1 & 335M & 50.4 & 67.3 & 73.1 & 80.4 & 94.4 & 99.4 & 59.2 & 65.1 \\
 & BMRetriever-410M & 410M & 60.6 & 72.9 & 76.5 & 82.2 & 95.0 & 99.1 & 66.7 & 70.9 \\
 \bottomrule
\end{tabular}
}
\caption{Full retrieval results across state of the art retrievers. 
Values are in percentage. R@k denotes recall at k.}\label{tab:retrieval-table}
\end{table}

\subsection{Full RAG Result}\label{app:full_rag}
\newcommand{\lighthl}[1]{\colorbox{yellow!20}{#1}}
\begin{table*}[ht]
\small
\centering
\setlength{\tabcolsep}{2.5pt}
\begin{tabular}{c l c c c c c c c}
\toprule
\multicolumn{1}{c}{\textbf{\# Reports}} & \multicolumn{1}{c}{\textbf{Model}} & \multicolumn{1}{c}{\textbf{BM25}} & \multicolumn{1}{c}{\textbf{\(\Delta\) BM25}} & \multicolumn{1}{c}{\textbf{BGE}} & \multicolumn{1}{c}{\textbf{\(\Delta\) BGE}} & \multicolumn{1}{c}{\textbf{BMRet.}} & \multicolumn{1}{c}{\textbf{\(\Delta\) BMRet.}} & \multicolumn{1}{c}{\textbf{Oracle}} \\
\midrule
\multirow{9}{*}{1} & GPT-5.2 & {77.0} & 21.0 & {76.8} & 20.8 & {74.8} & 18.8 & 89.8 \\
 & GPT-o4-mini & 73.6 & 21.8 & 73.2 & 21.4 & 72.5 & 20.7 & {84.4} \\
 & Llama 3.3 70B Instruct & 64.9 & 20.0 & 69.2 & 24.3 & 63.8 & 18.9 & {82.5} \\
 & Claude 4.5 Sonnet & 71.4 & 20.2 & 71.0 & 19.9 & 70.0 & 18.8 & {84.4} \\
 & Thinking Claude 4 Sonnet & 69.3 & 21.1 & 71.0 & 22.8 & 70.9 & 22.7 & {86.7} \\
 & Gemini 2.5 Pro & 68.5 & 21.9 & 71.8 & 25.2 & 69.0 & 22.4 & {86.5} \\
 & MedGemma-27b-text-it & 59.0 & 21.9 & 63.4 & {26.3} & 60.6 & {23.5} & {78.7} \\
 & Llama 3-Med42-70B & 60.7 & 18.6 & 63.4 & 21.3 & 59.9 & 17.8 & {78.1} \\
 & OpenBioLLM-Llama 3-70B & 55.9 & 18.3 & 63.1 & 25.5 & 58.4 & 20.8 & {76.1} \\
 \midrule
\multirow{9}{*}{3} & GPT-5.2 & 81.2 & 25.2 & 81.1 & 25.0 & 77.6 & 21.6 & - \\
 & GPT-o4-mini & 77.3 & 25.5 & 77.8 & 26.0 & 76.1 & 24.3 & - \\
 & Llama 3.3 70B Instruct & 67.6 & 22.7 & 69.5 & 24.6 & 69.2 & 24.3 & - \\
 & Claude 4.5 Sonnet & 71.7 & 20.5 & 73.7 & 22.5 & 70.6 & 19.4 & - \\
 & Thinking Claude 4 Sonnet & 75.4 & 27.2 & 76.1 & 27.9 & 72.8 & 24.6 & - \\
 & Gemini 2.5 Pro & 73.4 & 26.8 & 74.3 & 27.7 & 73.7 & 27.1 & - \\
 & MedGemma-27b-text-it & 62.0 & 24.9 & 64.0 & 26.9 & 62.8 & 25.7 & - \\
 & Llama 3-Med42-70B & 55.7 & 13.6 & 59.2 & 17.1 & 56.8 & 14.7 & - \\
 & OpenBioLLM-Llama 3-70B & 43.7 & 6.1 & 47.6 & 10.0 & 47.7 & 10.2 & - \\
 \midrule
\multirow{7}{*}{5} & GPT-5.2 & 80.1 & 24.1 & 82.2 & 26.1 & 79.0 & 23.0 & - \\
 & GPT-o4-mini & 77.5 & 25.7 & 77.2 & 25.4 & 76.4 & 24.6 & - \\
 & Llama 3.3 70B Instruct & 70.7 & 25.8 & 72.6 & 27.7 & 68.9 & 23.9 & - \\
 & Claude 4.5 Sonnet & 73.4 & 22.2 & 74.0 & 22.8 & 71.5 & 20.3 & - \\
 & Thinking Claude 4 Sonnet & 73.2 & 25.0 & 76.7 & 28.5 & 72.5 & 24.3 & - \\
 & Gemini 2.5 Pro & 73.2 & 26.6 & 76.7 & 30.0 & 72.6 & 26.0 & - \\
 & MedGemma-27b-text-it & 50.7 & 13.6 & 59.8 & 22.7 & 57.3 & 20.2 & - \\
 \bottomrule
\end{tabular}
\caption{Performance of RAG with different retrieval methods and context length. $\Delta$ stands for the percentage improvement from baseline. For retrieval with BGE, BGE-large-en-v1.5 was used.}
\label{tab:rag_results}
\end{table*}

\subsection{Retrieval Method Exploration}

Given the long lengths of our corpus and queries, along with the context-window limitations summarized in \autoref{tab:tokennum} and \autoref{tab:maxctx}, we employ a text processing strategy to optimize document representation. Documents are chunked with a maximum length of 512 tokens and a stride of 128. We then aggregate passage-level scores using a two-level Maximum Passage (MaxP) strategy \citep{Dai_2019}. \sk{this whole paragraph can also be moved to the appendix and just say, details of chunking etc for retrieval are in the appendix.}
\begin{table*}[h]
\centering

\begin{subtable}{0.48\textwidth}
    \centering

\centering
\small
\begin{tabular}{lrr}
\toprule
\textbf{Category} & \textbf{Avg. tokens} & \textbf{Max tokens} \\
\midrule
Corpus & 2730.21 & 24271 \\
Question  & 480.64  & 771   \\
Answer & 29.74   & 95   \\
\bottomrule
\end{tabular}
\caption{Token statistics for the retrieval corpus and queries and answers from \benchmark (Computed using Contriever's tokenizer \citep{izacard2022unsuperviseddenseinformationretrieval}).\sk{could move this to appendix also or just incorporate in text.}}
\label{tab:tokennum}
\end{subtable}
\hfill
\begin{subtable}{0.48\textwidth}
    \centering
\centering

\small
\setlength{\tabcolsep}{4pt}
\begin{tabular}{lr} 
\toprule
\textbf{Model} & \textbf{Context Length} \\
\midrule
GPT-5.2                       & 400K \\
GPT-o4-mini                 & 200K \\
Llama 3.3 70B Instruct & 128K \\
Claude 4.5 Sonnet   & 1M   \\
Thinking Claude 4 Sonnet & 1M \\
Gemini 2.5 Pro & 1M \\
MedGemma-27b-text-it        & 128K \\
Llama 3-Med42-70B            & 8K   \\
OpenBioLLM-Llama 3-70B       & 8K   \\
\bottomrule
\end{tabular}
\caption{Maximum context lengths of a subset of models we evaluate with RAG.}
\label{tab:maxctx}
\vspace{-0.4\baselineskip}
\end{subtable}
\caption{Overview of token length (left) and model context limits (right) guide the retrieval processing strategy.}
\end{table*}

\subsubsection{Comparison of Chunking Methods}\label{app:chunking}
\begin{table*}[h]
\resizebox{\textwidth}{!}{ 
\begin{tabular}{lllrrrrrrrr}
\toprule
\textbf{Model} & \textbf{Corpus} & \textbf{Query} & \multicolumn{1}{l}{\textbf{Recall@1}} & \multicolumn{1}{l}{\textbf{Recall@3}} & \multicolumn{1}{l}{\textbf{Recall@5}} & \multicolumn{1}{l}{\textbf{Recall@10}} & \multicolumn{1}{l}{\textbf{Recall@100}} & \multicolumn{1}{l}{\textbf{Recall@1000}} & \multicolumn{1}{l}{\textbf{MRR@5}} & \multicolumn{1}{l}{\textbf{nDCG@10}} \\
\midrule
\multirow{3}{*}{BGE-large-en-v1.5} & truncation & truncation & 35.2 & 50.2 & 57.9 & 66.0 & 87.6 & 96.9 & 43.5 & 49.7 \\
 & chunk & truncation & 62.9 & 76.4 & 81.5 & 86.7 & 96.1 & \textbf{99.4 }& 70.2 & 74.7 \\
 & chunk & chunk & \textbf{64.3} & \textbf{78.1} & \textbf{82.9} & \textbf{87.5} & \textbf{96.4} &\textbf{ 99.4} & \textbf{71.6} & \textbf{75.9} \\
 \midrule
\multirow{3}{*}{BMRetriever-410M} & truncation & truncation & 32.2 & 48.4 & 52.9 & 59.9 & 82.5 & 95.1 & 40.4 & 45.8 \\
 & chunk & truncation & \textbf{60.6} & 72.8 & \textbf{76.8} & \textbf{82.3} & \textbf{95.0} & \textbf{99.1} & \textbf{66.7} & \textbf{71.0} \\
 & chunk & chunk & \textbf{60.6} & \textbf{72.9} & 76.5 & 82.2 & \textbf{95.0} & \textbf{99.1} & \textbf{66.7} & 70.9 \\
 \bottomrule
\end{tabular}
}
\caption{Retrieval results using different chunking methods.}
\label{tab:chunk-vs-trunc}
\end{table*}
\autoref{tab:chunk-vs-trunc} presents the results of our chunking experiments on two top-performing retrievers, BGE-large and BMRetriever. We compared different combinations of chunking and truncation. Our chunking strategy used a maximum length of 512 tokens with a stride of 128, while truncation was a simple cut-off at 512 tokens. The results demonstrate that applying chunking to both the corpus and the query is essential for achieving high performance in our use case.
\subsubsection{Effects of context length}\label{app:context len}
\begin{table}[ht]
\centering
\resizebox{\textwidth}{!}{
\setlength{\tabcolsep}{4pt}
\scriptsize
\begin{tabular}{l c r r r r r r r r}
\toprule
\textbf{Model} & \textbf{Max Len} & \textbf{Recall@1} & \textbf{Recall@3} & \textbf{Recall@5} & \textbf{Recall@10} & \textbf{Recall@100} & \textbf{Recall@1000} & \textbf{MRR@5} & \textbf{nDCG@10} \\
\midrule
\multirow{2}{*}{BMRetriever-410M}
& 512 & \textbf{60.6}&	\textbf{72.9}&	\textbf{76.5}&	\textbf{82.2}&	\textbf{95.0}&	\textbf{99.1}&	\textbf{66.7}&	\textbf{70.9} \\

& 1024 &48.2&	62.1&	67.3&	72.0&	88.9&	97.8&	55.6&	60.1\\
\bottomrule
\end{tabular}
    }
\caption{Retrieval results using different context lengths.}
\label{tab:ctxlen}
\end{table}
To evaluate the impact of context length, we conducted an experiment with BMRetriever, which supports a maximum context length of 1,024 tokens and was tested with a fixed stride of 128. The results, presented in \autoref{tab:ctxlen}, indicate that merely increasing the context window does not necessarily yield improved performance—particularly for long, domain-specific medical queries such as those in our dataset.

\subsubsection{Effect of stride values}\label{app:strides}
\begin{table*}[h]
\resizebox{\textwidth}{!}{ 
\begin{tabular}{lcrrrrrrrr}
\toprule
\textbf{Model} & \textbf{Stride} & \multicolumn{1}{l}{\textbf{Recall@1}} & \multicolumn{1}{l}{\textbf{Recall@3}} & \multicolumn{1}{l}{\textbf{Recall@5}} & \multicolumn{1}{l}{\textbf{Recall@10}} & \multicolumn{1}{l}{\textbf{Recall@100}} & \multicolumn{1}{l}{\textbf{Recall@1000}} & \multicolumn{1}{l}{\textbf{MRR@5}} & \multicolumn{1}{l}{\textbf{nDCG@10}} \\
\midrule
\multirow{4}{*}{BGE-large-en-v1.5} & 128 & \textbf{64.3} & \textbf{78.1} & \textbf{82.9} & \textbf{87.5} & \textbf{96.4} & \textbf{99.4} & \textbf{71.6} & \textbf{75.9} \\
 & 256 & 57.6 & 71.8 & 76.8 & 84.2 & 95.0 & \textbf{99.4} & 64.9 & 70.3 \\

 & 384 & 56.7 & 69.8 & 75.3 & 81.1 & 95.1 & \textbf{99.4} & 63.7 & 68.5 \\
&512 & 50.5 & 64.3 & 70.1 & 77.6 & 94.1 & \textbf{99.4} & 58.1 & 63.5 \\
 \midrule
\multirow{4}{*}{BMRetriever-410M} & 128 &\textbf{60.6} & \textbf{72.9} & \textbf{76.5} & \textbf{82.2} &\textbf{ 95.0} &\textbf{ 99.1} & \textbf{66.7} &\textbf{ 70.9} \\

 & 256 & 51.8 & 65.6 & 70.0 & 75.9 & 92.6 & 98.9 & 59.0 & 63.7 \\

 & 384 & 48.7 & 62.0 & 67.1 & 74.0 & 91.7 & 98.6 & 55.7 & 60.8 \\

 &512 & 43.0 & 56.7 & 63.5 & 69.8 & 89.4 & 98.0 & 50.5 & 55.8\\
 \bottomrule
\end{tabular}
}
\caption{Retrieval results using different stride values.}
\label{tab:stride}
\end{table*}
To see the effect of different stride values, we conducted experiments on the two top-performing retrievers, BGE-large and BMRetriever models, with a fixed maximum context length of 512. The results, as detailed in \autoref{tab:stride}, revealed that a stride of 128 consistently outperformed other configurations. Consequently, this stride value was selected for all subsequent experiments.

\subsubsection{Retrieval Result Under Simple Truncation}\label{app:retrieval truncation}
\begin{table*}[h]
\centering

\setlength{\tabcolsep}{1.5pt}   
\small
\resizebox{\textwidth}{!}{    
\begin{tabular}{clcrrrrrrrr}
\toprule
\multicolumn{1}{c}{\textbf{Type}} & \multicolumn{1}{c}{\textbf{Model}} & \multicolumn{1}{c}{\textbf{Params}} & \multicolumn{1}{c}{\textbf{R@1}} & \multicolumn{1}{c}{\textbf{R@3}} & \multicolumn{1}{c}{\textbf{R@5}} & \multicolumn{1}{c}{\textbf{R@10}} & \multicolumn{1}{c}{\textbf{R@100}} & \multicolumn{1}{c}{\textbf{R@1000}} & \multicolumn{1}{c}{\textbf{MRR@5}} & \multicolumn{1}{c}{\textbf{nDCG@10}} \\ \midrule
\multirow{8}{*}{General}& All-MiniLM-L12-v2 & 33.4M & 5.8 & 10.3 & 12.4 & 17.4 & 37.6 & 70.0 & 8.2 & 10.8 \\
 & E5-small-v2 & 33.4M & 15.6 & 26.4 & 31.6 & 38.3 & 66.0 & 87.9 & 21.3 & 26.0 \\
 & Contriever-msmarco & 110M & 15.2 & 23.5 & 28.6 & 34.9 & 63.7 & 87.0 & 20.0 & 24.2 \\
 & Contriever & 110M & 16.9 & 23.0 & 26.4 & 30.4 & 53.1 & 78.6 & 20.5 & 23.2 \\
 & BGE-small-en-v1.5 & 33.4M & 30.2 & 45.1 & 49.5 & 56.7 & 82.8 & 96.1 & 37.6 & 42.9 \\
 & BGE-base-en-v1.5 & 109M & 30.2 & 42.6 & 49.3 & 58.4 & 83.1 & 96.4 & 37.1 & 43.1 \\
 & BGE-large-en-v1.5 & 335M & 35.2 & 50.2 & 57.9 & 66.0 & 87.6 & 96.9 & 43.5 & 49.7 \\
 & BGE-m3 & 560M & 22.2 & 32.7 & 37.6 & 43.7 & 68.2 & 88.3 & 28.0 & 32.3 \\

 \midrule
\multirow{6}{*}{Finetuned} & MedCPT-Query-Encoder & 109M & 18.0 & 28.2 & 33.0 & 42.7 & 71.8 & 92.3 & 23.4 & 28.9 \\

 & Pubmedbert-base-embeddings & 109M & 24.7 & 37.6 & 45.4 & 53.2 & 81.5 & 95.9 & 32.2 & 38.0 \\

 & MedEmbed-small-v0.1 & 33.4M & 27.4 & 38.7 & 44.9 & 54.0 & 81.5 & 96.1 & 33.7 & 39.5 \\

 & MedEmbed-base-v0.1 & 109M & 29.6 & 43.0 & 47.7 & 56.3 & 82.0 & 96.9 & 36.4 & 42.0 \\

 & MedEmbed-large-v0.1 & 335M & 29.1 & 44.3 & 50.4 & 58.1 & 84.8 & 96.7 & 37.2 & 42.9 \\

 & BMRetriever-410M & 410M & 32.2 & 48.4 & 52.9 & 59.9 & 82.5 & 95.1 & 40.4 & 45.8 \\

 \bottomrule
\end{tabular}
}
\caption{Retrieval results using simple truncation. Values are in percentage. R@k
denotes recall at k.}
\label{tab:retrieval-truncation}
\end{table*}
\autoref{tab:retrieval-truncation} reports retrieval performance under a simple truncation strategy with a maximum context length of 512 tokens for both corpus and queries. As expected, performance is consistently lower than with the chunking strategy, with no model achieving Recall@1 above 50\%. This highlights the importance of chunking and reveal substantial room for improvement in modern retrievers, particularly for rare-case retrieval.

\subsubsection{Retrieval Results Using Reranker}
\newcommand{\pos}[1]{\cellcolor{green!15}#1}
\newcommand{\negc}[1]{\cellcolor{red!15}#1}


\begin{table}[h]
\small
\centering
\begin{tabular}{l rr rr}
\toprule
\multirow{2}{*}{\textbf{model}} 
& \multicolumn{2}{c}{\textbf{MedCPT-Cross-Encoder}} 
& \multicolumn{2}{c}{\textbf{BGE-reranker-large}} \\
\cmidrule(lr){2-3} \cmidrule(lr){4-5}
& \textbf{Recall@10$\uparrow$} & \textbf{NDCG@10$\uparrow$} 
& \textbf{Recall@10$\uparrow$} & \textbf{NDCG@10$\uparrow$} \\
\midrule
All-MiniLM-L12-v2 & \pos{61.3} & \pos{115.4} & \negc{-8.4} & \pos{3.5} \\
E5-small-v2 & \pos{13.8} & \pos{23.1} & \negc{-19.2} & \negc{-25.9} \\
Contriever-msmarco & \pos{18.2} & \pos{26.9} & \negc{-12.6} & \negc{-17.9} \\
Contriever & \pos{27.8} & \pos{34.2} & \negc{-0.8} & \negc{-6.0} \\
BGE-small-en-v1.5 & \pos{10.1} & \pos{14.2} & \negc{-33.3} & \negc{-39.7} \\
BGE-base-en-v1.5 & \pos{2.1} & \pos{6.7} & \negc{-35.5} & \negc{-42.1} \\
BGE-large-en-v1.5 & \pos{0.2} & \negc{-3.2} & \negc{-38.6} & \negc{-48.8} \\
BGE-m3 & \pos{11.3} & \pos{16.5} & \negc{-30.0} & \negc{-35.9} \\
\midrule
MedCPT-Query-Encoder & \pos{26.8} & \pos{48.7} & \negc{-10.1} & \negc{-6.5} \\
Pubmedbert-base-embeddings & \pos{14.5} & \pos{25.4} & \negc{-29.0} & \negc{-31.4} \\
MedEmbed-small-v0.1 & \pos{15.5} & \pos{30.5} & \negc{-29.1} & \negc{-32.9} \\
MedEmbed-base-v0.1 & \pos{6.5} & \pos{13.5} & \negc{-40.4} & \negc{-47.4} \\
MedEmbed-large-v0.1 & \pos{5.1} & \pos{9.6} & \negc{-42.8} & \negc{-50.3} \\
BMRetriever-410M & \pos{7.4} & \pos{5.2} & \negc{-36.8} & \negc{-48.0} \\ \bottomrule
\end{tabular}
\caption{Percentage improvement in retrieval using MedCPT-Cross-Encoder and BGE-reranker-large.}
\label{tab:ce-medcpt}
\end{table}

As shown in \autoref{tab:ce-medcpt}, incorporating a strong biomedical reranker, MedCPT-Cross-Encoder, consistently improves the performance of most retrieval models. However, this improvement varies greatly according to retriever models, and several larger retriever models do not have a notable difference, suggesting that simply adding a specialized reranker is not always a guaranteed solution. BGE-reranker-large Meanwhile, \citep{xiao2024cpackpackedresourcesgeneral} exhibits a notable decline in performance across all retriever models. This result highlights that even state-of-the-art rerankers struggle to perform effective reranking within the context of \benchmark{}.

\subsection{LLM-as-a-Judge}\label{app:llm-as-a-judge}
Using the Wilson confidence interval, the 95\% confidence interval for expert agreement of our LLM-as-a-judge evaluation is 85.6\% to 96.9\%. This surpasses the agreement levels in MT-Bench\citep{zheng2023judgingllmasajudgemtbenchchatbot}, an extremely popular benchmark, which has human-human agreement of approximately 81\% and LLM-human agreement of above 80\%.

\subsection{Bootstrap Analysis}\label{app:bootstrap}
\begin{table}[h]
\centering

\begin{subtable}{\textwidth}
\centering
\small
\begin{tabular}{lccc}
\toprule
\textbf{Model} & \textbf{Accuracy} & \textbf{Mean} & \textbf{95\% CI} \\
\midrule
GPT-5.2 & 56.0 & 55.97 & 52.27 - 59.94 \\
GPT-o4-mini & 51.8 & 51.81 & 47.73 - 55.71 \\
MedGemma-27b-text-it & 37.1 & 37.09 & 33.18 - 41.00 \\
Llama 3-Med42-70B & 42.1 & 42.05 & 38.34 - 45.70 \\
\bottomrule
\end{tabular}
\caption{Baseline}
\end{subtable}

\vspace{8pt}
\begin{subtable}{\textwidth}
\centering
\small
\begin{tabular}{lccc}
\toprule
\textbf{Model} & \textbf{Accuracy} & \textbf{Mean} & \textbf{95\% CI} \\
\midrule
GPT-5.2 & 81.1 & 81.04 & 78.09 - 84.19 \\
GPT-o4-mini & 77.8 & 77.80 & 74.80 - 80.91 \\
MedGemma-27b-text-it & 64.0 & 64.01 & 60.25 - 67.76 \\
Llama 3-Med42-70B & 59.2 & 59.14 & 55.24 - 62.76 \\
\bottomrule
\end{tabular}
\caption{RAG (BGE-large, top-3)}
\end{subtable}

\caption{Bootstrap analysis of two representative setups.}
\label{tab:bootstrap}
\end{table}
We used bootstrap analysis, which is a resampling method of repeatedly drawing samples with replacement to approximate the sampling distribution, to confirm that our benchmark shows stable results. 1,000 bootstrap sampling of both the baseline and BGE retriever with 3 reports as context in \autoref{tab:bootstrap} confirms that our benchmark has a stable performance estimate. The bootstrap mean is nearly identical to the reported accuracy. This shows that our clinically rich benchmark of 639 questions is sufficient for model comparison.

\newpage

\section{Example Data Creation}

\begin{figure*}[h]
\centering
\begin{tcolorbox}[txtbox, width=\textwidth]
\small 
\RaggedRight

\textbf{Bleeding anorectal varices treated by a direct puncture approach through the greater sciatic foramen: The utility of a steerable microcatheter for reverse catheterization}\\
PMCID: PMC8829532 \hfill Internal Medicine

\medskip
\textbf{I. INTRODUCTION}\\
Anorectal varices represent a developed portosystemic collateral pathway where portal venous blood flows from the superior rectal veins of the inferior mesenteric system to the middle and inferior rectal veins of the iliac system. Anorectal varices most commonly result from portal hypertension \ldots\ (Omitted for brevity).

\medskip
\textbf{II. CASE PRESENTATION}\\
\textbf{The patient was a woman in her 70s with Child-Pugh class B primary biliary cholangitis. She was admitted to our hospital for treatment of intractable anorectal varices. Her gastroesophageal varices were repeatedly treated at another hospital, 5 and 2 years ago, respectively} \ldots\ (Omitted for brevity).

\medskip
\textbf{III. DISUCSSION}\\
\ldots\ (Omitted for brevity) It can be managed by using a variety of methods, including endoscopic therapy (EIS and EBL), IR approach, and surgical treatment, but no standard treatment has been established yet. \textcolor{red}{In our case, large, dilated varices rendered both endoscopic options unsuitable. EIS was considered dangerous because, given the high blood flow \ldots\ (Omitted for brevity) \ldots\ TIPS and PTO cannot be conducted in cases, like ours, where ascites and portal vein thrombosis are present.} \textcolor{blue}{Direct puncture access to the superior rectal vein approached through the greater sciatic foramen was pioneered by Kariya et al., who first used this pelvic approach for successful endovascular treatment for rectal varices and reported its efficacy and feasibility.} \ldots\ (Omitted for brevity).

\medskip
\textbf{QUESTION}\\
A female patient in her eighth decade with moderate hepatic dysfunction secondary to primary biliary cirrhosis presents with life-threatening hemorrhoidal variceal bleeding.  \ldots\ (Omitted for brevity) \ldots\
feeding vessel to the perianal varices, along with moderate volume ascitic fluid. \textcolor{red}{Traditional interventional approaches including balloon-assisted venous occlusion (complicated by numerous outflow vessels), transjugular portosystemic shunting and percutaneous liver-directed obliteration (both contraindicated by portal thrombosis and peritoneal fluid) are all excluded.} \ldots\ (Omitted for brevity) \ldots\ given the limitations precluding standard endoscopic or conventional radiologic therapies, \textit{what emergent procedural intervention would provide optimal hemorrhage control?}

\medskip
\textbf{ANSWER}\\
Proceed with variceal obliteration via \textcolor{blue}{direct puncture access to the dilated right superior rectal vein}, approaching it through the left greater sciatic foramen.

\end{tcolorbox}

\caption{Example of case report and corresponding final question-answer pair. Timeline is \textbf{bolded}, significance is in \textcolor{blue}{blue}, and limitation is in \textcolor{red}{red}. Portions of original contents have been selectively omitted for brevity and irrelevance. The question asks the direct next step given the patient details (\textit{italicized}), and the answer is related to the significant point of the case report. Link for full text: \url{https://pmc.ncbi.nlm.nih.gov/articles/PMC8829532/}.}
\label{fig:case_report_excerpt}
\end{figure*}

\newpage
\section{Failure Mode}\label{app:failure-mode-full}
\subsection{Results}\label{app:failure_mode}
\vspace{-2.0em}
\begin{figure}[h]
\centering
\includegraphics[
    width=0.98\linewidth,
    trim=30pt 90pt 45pt 14pt,
    clip
]{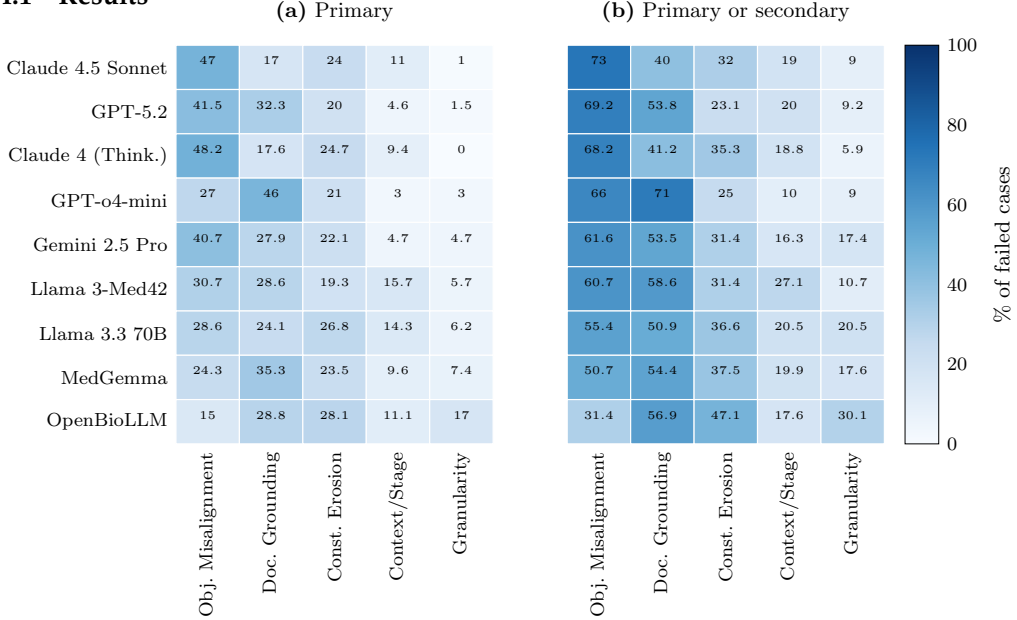}
\caption{\textbf{Failure mode distribution on \benchmark.}
Cells show the percentage of failed cases for which each mode is primary (left)
or primary/secondary (right).}
\label{fig:failure-modes-combined}
\vspace{-0.8em}
\end{figure}
\vspace{-0.5em}
\subsection{Failure Mode Illustrative Examples}

\begin{figure}[h]
\centering
\begin{tcolorbox}[txtbox, width=\textwidth]
\small
\RaggedRight

\textbf{Acute myocardial infarction following radiofrequency catheter ablation in a child}\\
PMCID: PMC11049577 \hfill Internal Medicine

\smallskip
\textbf{Case summary}\\
A pre-adolescent female underwent catheter ablation for drug-refractory AVNRT.
Post-procedurally, inferior ST-segment elevations developed; emergency coronary angiography
confirmed right coronary occlusion at the AV-nodal artery.
Repeated vasodilator administration failed; patency was restored by wire manipulation,
and \uline{intravascular imaging was performed at that time}.
The team deferred stenting given the patient's age. With \uwave{persistent uncertainty
about the mechanism of obstruction, and intravascular imaging already completed},
further characterization was sought.

\smallskip
\textbf{Question}\\
\ldots\ With persistent uncertainty about the pathophysiology despite restored flow,
\textit{what diagnostic modality would best characterize the vascular abnormality?}

\smallskip
\textbf{Oracle answer}\\
\uline{Obtain coronary computed tomography angiography (CCTA) for further evaluation.}

\smallskip
\textbf{Model output}\\
Intravascular ultrasound (IVUS) allows detailed visualization of the vessel wall \ldots\
clarifying whether the stenosis is due to direct thermal injury or extrinsic compression.

\smallskip
\textit{The article specifies CCTA as the next step after intravascular imaging has already been
performed. The model recommends IVUS, the modality the case report states was already
completed, failing to stay anchored to the document-specific investigative sequence.}
\end{tcolorbox}
\caption{\textbf{Document grounding failure.}
The model substitutes a previously completed diagnostic modality (IVUS)
for the article-specified next investigation (CCTA),
indicating insufficient grounding to the oracle document's stated clinical course.
\uline{Single underlines} mark content directly informing the oracle answer; \uwave{wavy underlines} mark constraining clinical context the model disregarded.}
\label{fig:fm-doc-grounding}
\end{figure}

\vspace{-5em}

\begin{figure}[htbp]
\centering
\begin{tcolorbox}[txtbox, width=\textwidth]
\small
\RaggedRight

\textbf{Late-onset bleeding from an intercostal artery pseudoaneurysm after hemostasis using soft coagulation}\\
PMCID: PMC10899919 \hfill Surgical Studies

\smallskip
\textbf{Case summary}\\
A 68-year-old male who previously underwent right lower lobe lobectomy and a
subsequent wedge resection presented emergently with hemoptysis and severe hypoxemia
(SpO\textsubscript{2} 70\%). Contrast-enhanced CT demonstrated an 11~cm
right hemothorax with \uline{active extravasation from a pseudoaneurysm of the
sixth intercostal artery}.
Three-dimensional vascular reconstruction confirmed the false aneurysm.

\smallskip
\textbf{Question}\\
\ldots\ Given active vascular extravasation from an intercostal arterial false aneurysm
causing hemodynamic compromise and severe hypoxemia,
\textit{what intervention should be pursued urgently?}

\smallskip
\textbf{Oracle answer}\\
\uline{Arrange immediate emergency angiography through the
Radiology/Interventional Radiology service.}

\smallskip
\textbf{Model output}\\
The patient should undergo urgent angiography \textbf{with embolization}
to stop the bleeding from the intercostal arterial false aneurysm.

\smallskip
\textit{The oracle specifies the immediate next action: arranging angiography via IR.
The model correctly identifies this step but appends embolization, a subsequent therapeutic
decision that follows angiographic confirmation, thereby answering a broader task
(the full interventional procedure) rather than the designated next action.}
\end{tcolorbox}
\caption{\textbf{Objective misalignment failure.}
The model includes the oracle-specified step but expands its answer to encompass a
downstream intervention not designated by the article,
reflecting optimization toward clinical completeness rather than oracle fidelity.
\uline{Single underlines} mark content directly informing the oracle answer.}
\label{fig:fm-obj-misalign}
\end{figure}

\begin{figure}[h]
\centering
\begin{tcolorbox}[txtbox, width=\textwidth]
\small
\RaggedRight

\textbf{Conversion surgery for advanced jejunal adenocarcinoma with multiple peritoneal metastases}\\
PMCID: PMC10435430 \hfill Surgical Studies

\smallskip
\textbf{Case summary}\\
A 64-year-old male with Stage~IV jejunal adenocarcinoma and peritoneal carcinomatosis
achieved a near-complete response after 4.5~years of systemic therapy, with laparoscopic
confirmation of absent peritoneal disease.
He now presents with \uline{chronic hemorrhage from the prosthesis site} causing
transfusion-dependent iron-deficiency anemia, and deteriorating performance status
\uwave{precluding further cytotoxic therapy}. Exploratory laparoscopy confirmed no
residual abdominal metastases.

\smallskip
\textbf{Question}\\
\ldots\ Given uncontrolled prosthesis-related blood loss and chemotherapy intolerance
following laparoscopic confirmation of absent metastases,
\textit{what intervention should be pursued?}

\smallskip
\textbf{Oracle answer}\\
\uline{Proceed with partial jejunectomy.}

\smallskip
\textbf{Model output}\\
The intervention that should be pursued in this case is conversion surgery.

\smallskip
\textit{Both answers identify surgery as the required intervention class.
The model uses the category-level label (conversion surgery) without naming
the specific operative procedure. The oracle requires the exact operation (partial jejunectomy);
the model's answer is correct in type but insufficiently specified to be equivalent.}
\end{tcolorbox}
\caption{\textbf{Granularity mismatch failure.}
The model correctly identifies the intervention class (surgery) but answers at a
higher level of abstraction than the oracle requires, naming the strategic category
rather than the specific procedure.
\uline{Single underlines} mark content directly informing the oracle answer; \uwave{wavy underlines} mark constraining clinical context the model disregarded.}
\label{fig:fm-granularity}
\end{figure}

\begin{figure}[h]
\centering
\begin{tcolorbox}[txtbox, width=\textwidth]
\small
\RaggedRight

\textbf{Use of the Stingray Re-Entry System in two complex cases of occluded superficial femoral arteries}\\
PMCID: PMC3420382 \hfill Internal Medicine

\smallskip
\textbf{Case summary}\\
A 77-year-old female with Rutherford category~3 PAD and complete proximal right SFA
occlusion underwent percutaneous revascularization.
A CrossBoss device was advanced to the popliteal level via a subintimal route,
then exchanged for the Stingray Re-Entry System over a Miracle guidewire.
\uline{The Stingray catheter is now positioned subintimally at the popliteal artery
following device exchange.}

\smallskip
\textbf{Question}\\
\ldots\ With the wire and catheter positioned subintimally after device exchange to the
Stingray system,
\textit{what represents the most suitable immediate subsequent procedural maneuver?}

\smallskip
\textbf{Oracle answer}\\
\uline{Inflate the Stingray Re-Entry System balloon to stabilize the catheter
within the subintimal space at the popliteal artery.}

\smallskip
\textbf{Model output}\\
Advance the dedicated Stingray Re-Entry System guidewire through the catheter's
distal port to penetrate the intima into the true lumen.

\smallskip
\textit{The article defines two sequential steps: (1) balloon inflation for subintimal
stabilization, followed by (2) guidewire advancement for true-lumen re-entry.
The question asks for step~1; the model provides step~2,
selecting an action from the immediately subsequent procedural stage.}
\end{tcolorbox}
\caption{\textbf{Context/stage misbinding failure.}
The model skips the current procedural step (balloon inflation for stabilization)
and provides the immediately subsequent maneuver (guidewire advancement),
selecting an action bound to the wrong point in the intervention sequence.
\uline{Single underlines} mark content directly informing the oracle answer.}
\label{fig:fm-stage}
\end{figure}

\begin{figure}[h]
\centering
\begin{tcolorbox}[txtbox, width=\textwidth]
\small
\RaggedRight

\textbf{Newly diagnosed AIDS patient with cerebellar JC virus}\\
PMCID: PMC10461121 \hfill Neurology

\smallskip
\textbf{Case summary}\\
A 57-year-old male with AIDS-related progressive multifocal leukoencephalopathy
(PML) achieved HIV suppression on antiretroviral therapy but continued to deteriorate
neurologically, with repeat CSF showing \uwave{JC virus burden rising from
2,000 to 14,000~copies/mL}.
Empiric glucocorticoids for suspected immune reconstitution syndrome provided no benefit.
He subsequently developed multi-domain neurological deficits including \uwave{new
confusion, dysphagia, language impairment, and increasing left-sided weakness}.

\smallskip
\textbf{Question}\\
\ldots\ Given ongoing neurological decline with expanding deficits despite sustained
HIV control and failure of glucocorticoids,
\textit{what immediate therapeutic intervention should be pursued?}

\smallskip
\textbf{Oracle answer}\\
\uline{Re-hospitalize him} and \uline{initiate pembrolizumab therapy},
\uline{with repeat neuroimaging to assess for progression.}

\smallskip
\textbf{Model output}\\
The immediate therapeutic intervention that should be pursued for this patient is
the initiation of pembrolizumab therapy.

\smallskip
\textit{The oracle specifies a three-component bundle: re-hospitalization,
pembrolizumab initiation, and repeat neuroimaging.
The model isolates the key therapeutic agent while dropping the organizational
(re-hospitalization) and monitoring (neuroimaging) components,
producing an answer that is related but not equivalent to the oracle.}
\end{tcolorbox}
\caption{\textbf{Constraint/qualifier erosion failure.}
The model identifies the correct therapeutic agent but omits two required concurrent
actions, producing an answer that is a strict subset of the oracle rather than equivalent to it.
\uline{Single underlines} mark content directly informing the oracle answer; \uwave{wavy underlines} mark constraining clinical context the model disregarded.}
\label{fig:fm-constraint}
\end{figure}

\newpage

\clearpage
\section{Prompts and Instructions} \label{app:prompts}

\begin{figure}[ht]
\centering
\begin{tcolorbox}[colframe=yellow!20!white, colback=yellow!5!white,
  boxrule=0.4pt,width=\textwidth,arc=1.5mm,auto outer arc,
  title=BMRetriever Instruction, fonttitle=\color{black}\bfseries,
  left=2pt,right=2pt,top=2pt,bottom=2pt]

\setstretch{1.0}\footnotesize
\textbf{Query template.} Given a question, retrieve relevant documents that best answer the question. \emph{\texttt{<query>}} \\
\textbf{Passage template.} Represent this passage\textbackslash n passage: \emph{\texttt{<passage>}}
\end{tcolorbox}
\caption{Instruction used by BMRetriever.}
\label{fig:bm_prompt_query}
\end{figure}

\begin{figure}[h!]
    \centering
    \begin{tcolorbox}[ colframe=pink!50!white, colback=pink!10!white, boxrule=0.5mm, width=\textwidth, arc=2mm, auto outer arc, title=GPT-5.2 prompt for significance extraction, fonttitle=\color{black}\bfseries]
    \setstretch{1.3}
Please carefully read the provided case report text or abstract. Identify whether the report describes any unique clinical actions from the following list:
\\[1em]
- Novel treatment or drug introduced

- Existing treatment used in a new way or indication

- New surgical or procedural technique applied

- Innovative combination of treatments or devices

- Novel diagnostic test or imaging method used

- Advanced molecular/genetic testing guiding treatment

- Unique point-of-care or biomarker test employed

- Use of AI or machine learning for diagnosis or treatment planning

- Novel intervention to manage unexpected treatment complications

- Off-label drug use with unique dosing or delivery method

- Personalized or precision medicine approach in therapy

- New integrated multidisciplinary care strategy

- Innovative use of telemedicine or remote monitoring in clinical management

- New rehabilitation or follow-up protocol applied

- Novel preventive or screening intervention implemented

- Ethical or legal decision-making impacting treatment

- Use of newly developed medical devices or technologies for treatment
\\[1em]
For each identified action, briefly summarize what it is and how it is unique or novel in this case report. If none apply, output exact string ``no" (without quote marks) only. 
If there are multiple points, only describe the MOST SIGNIFICANT POINT, most likely mentioned in the abstract. 
    
    \end{tcolorbox}
    \caption{GPT-5.2 Prompt we used to extract significance from the case reports. The output is used to create the question-answer pairs.}
    \label{fig:prompt_sig}
\end{figure}

\begin{figure}[h!]
    \centering
    \begin{tcolorbox}[colframe=pink!50!white, colback=pink!10!white, boxrule=0.5mm, width=\textwidth, arc=2mm, auto outer arc, title=GPT-5.2 prompt for timeline extraction, fonttitle=\color{black}\bfseries]
    \setstretch{1.3}
You will receive a clinical case presentation. Your task is to carefully parse and organize all clinical information strictly as a chronological timeline, listing events sequentially based on when they occurred or inferred from context.
Include ALL significant events: arrival, initial presentation, interventions, medications administered, diagnostic tests and results, progression events, clinical decisions, and final diagnosis.
\\[1em]
- Patient Arrival: Infer when and how the patient initially presented.

- Initial Clinical Presentation: Initial symptoms, vital signs, mental status, or general condition upon arrival.

- Early Medications and Interventions: Any initial treatments or interventions (medications, fluids, oxygen, etc.) performed upon or soon after presentation.

- Diagnostic Tests and Results: Tests ordered, procedures performed, and their outcomes. Infer order based on context if not explicitly stated.

- Clinical Progression: Any changes in the patient's condition over time, including improvements, deteriorations, or significant clinical events.

- Final Diagnosis or Impression: The final diagnosis, clinical assessment, or conclusion.
\\[1em]
Avoid categorizing by types of events; instead, present them sequentially as they unfold in real clinical practice.
Use exact sentences from the case report for building the timeline. 
DO NOT include any information that is not the a part of case presentation.
Within the case presentation, include any information necessary for clinical decision making, such as the patient refusing or requireing certain diagnostic tools. 
\\[1em]
Output should be formatted in similiar way as this for parsing purpose: 

- patient information 

- initial symptoms and patient state

- following timeline in chronological order

...

- final timeline of the case report 

    \end{tcolorbox}
    \caption{GPT-5.2 Prompt we used to extract timeline from the case reports. The output is used to create the question-answer pairs.}
    \label{fig:prompt_timeline}
\end{figure}

\begin{figure}[h!]
    \centering
    \begin{tcolorbox}[colframe=pink!50!white, colback=pink!10!white, boxrule=0.5mm, width=\textwidth, arc=2mm, auto outer arc, title=GPT-5.2 prompt for limitation extraction, fonttitle=\color{black}\bfseries]
    \setstretch{1.3}
You are given  (1) a medical case report, and (2) a paragraph describing the clinical significance or novelty of the case (i.e., why the case was written), referred to as "significance".
\\[1em]
Task:

Your task is to extract and summarize explicitly case-specific clinical or contextual limitations that prevented the use of standard or traditional management and motivated the use of novel approach described in the significance. 

A "limitation" must be an explicitly stated barrier to eligibility, access, feasibility, or safety of standard medical procedure. Disease severity, clinical difficulty, rarity, or the fact that the case is "hard" are NOT LIMITATIONS unless the case report explicitly states that they prevented use of standard management.

Prior treatment failure counts as a limitation only if explicitly stated as a reason standard management could not be used.

Limitation is stated in the Discussion section of the case report, and is different from the section header "limitation".
\\[1em]
Some case reports do not have limitations specifically stated. 
\\[1em]
Limitations may include (but not restricted to):

 - patient-specific factors (e.g., age, frailty, comorbidities, contraindications)
 
 - system-level constraints (e.g., limited resources, emergency, long wait times)
 
 - disease-specific constraints blocking standard medical procedure(e.g., anatomical complexity preventing surgery)
 
 - explicitly stated ineligibility for standard medical procedure
\\[1em]
The following are NOT limitations on their own:

 - disease severity
 
 - treatment difficulty
 
 - failed prior methods, unless stated as blocking standard care and reason why significance of the case report was used
 
 - general background or motivation of the case
 
 - novelty or importance of the case

Instructions:

 - identify limitations only if they are explicitly stated in the report (from the Discussion section)
 
 - DO NOT infer based on medical knowledge
 
 - DO NOT restate significance 
 
 - summarize only the reasons or constraints, not the procedure itself
 
 - if multiple limitations are mentioned, list all of them concisely
 
 - if no explicit limitation is stated, output exact string "no" (without quote marks) only

    \end{tcolorbox}
    \caption{GPT-5.2 Prompt we used to extract limitations from the case reports. The output is used to create the question-answer pairs.}
    \label{fig:prompt_limitation}
\end{figure}
\begin{figure}[h!]
    \centering
    \begin{tcolorbox}[colframe=pink!50!white, colback=pink!10!white, boxrule=0.5mm, width=\textwidth, arc=2mm, auto outer arc, title=GPT-5.2 prompt for question-answer pair generation, fonttitle=\color{black}\bfseries]
    \setstretch{1.3}

You are given a timeline and significant point of a medical case report. The significance refers to the new technique that the case report introduces. It can be surgical method, treatment, etc. Significance mentions the ``new" item and explains why it is new. Timeline is an ordered list of bullet-style events exactly as they appear in the case report. 
\\[1em]
Create one question-answer pair asking for the immediate next step at the moment just before the action happens. The question should be the steps before taking the ``new" action introduced in the significance. It should include all the steps before the decision point of ``new" action. If there are multiple significant points in the input, choose the most complicated and unique one. 
\\[1em]
The question should not be based on an outcome but asking what is the significance or next step in treatment, or whether certain treatment/test could be done for this specific case, what to keep in mind during the treatment, etc. It should not ask about the past.  
\\[1em]
Extract a concise answer, using only exact sentence(s) from the timeline. The answer must be verbatim, not paraphrased.
\\[1em]
This is a text only communication, so neither question nor answer should be referring to images, tables, or any other non-text media. 
\\[1em]
Output: 
\\[1em]
Context: Clearly list EVERY step of the timeline BEFORE this decision point in chronological order without leaving out any bulletpoints. ANY POINTS AFTER THE DECISION POINT SHOULD NOT BE IN THE CONTEXT. patient arrival, symptoms, vital signs, prior history, treatments/interventions, medications administered, diagnostic tests and their results, and clinical progression events—strictly up to this decision point should be included. If two consecutive elements of timeline are ``decided to do xxx" and ``xxx was performed", state it only once. 
\\[1em]
Question: Given this information, what is the immediate next appropriate clinical step?  
(Note: Adapt this to fit the situation, e.g., ``next appropriate test", ``next appropriate treatment", etc.)
\\[1em]
Answer: State the exact next clinical step taken, using precise procedural or diagnostic terminology. Only use information explicitly stated in the timeline. THIS SHOULD NOT BE INCLUDED IN THE END OF THE CONTEXT. THIS IS THE **NEXT** STEP AFTER THE QUESTION STATEMENT.
\\[1em]
The response should be outputted in json format. 
[Format omitted for brevity]
    \end{tcolorbox}
    \caption{GPT-5.2 Prompt we used to create question-answer pairs from the case reports. Output ``Context" and ``Question" are combined to form the question.}
    \label{fig:prompt_query}
\end{figure}
\begin{figure}[h!]
    \centering
    \begin{tcolorbox}[colframe=pink!50!white, colback=pink!10!white, boxrule=0.5mm, width=\textwidth, arc=2mm, auto outer arc, title=Claude 4 Opus prompt for controlled question modification, fonttitle=\color{black}\bfseries]
    \setstretch{1.3}
You are given three items:  
(1) a detailed clinical query,  
(2) its corresponding concise query,  
(3) the answer.
 \\[1em]
Your task is to rewrite the **detailed query** following the instructions below:
\\[1em]
Instructions:

1. Identify all overlapping content between the detailed and concise queries.

2. You must **preserve** the meaning of all overlapping content exactly yet **modify** the words or expressions. 

    - Keep the core meaning **unchanged**, but vary the surface form:  
    
    - Use synonyms, abbreviations, or different phrasing.  
    
    - Do NOT alter the medical intent or expected answer.  
    
    - Example: "Management of acute MI" → "Initial treatment of a heart attack".
    
3. Identify the non-overlapping parts of the detailed query:

    - Use synonyms, abbreviations, or different phrasing. 
    
    - Adjust numerical values by adding or subtracting within medically reasonable ranges.  
    
    - Altering the logical flow, sentence structure, or clinical context.
    
4. Add **extra distracting medical content** that are medically plausible but irrelevant to the answer:  

    - Comorbidities
    
    - Symptoms, tests, and treatments.  

    - Background information.  
    
    - Past but resolved medical history.  
    
    - Family history that does not affect the answer.  
    
    - Redundant or vague phrases.  
\\[1em]
**IMPORTANT:**  

- The revised detailed query should look **substantially different** from the original, while remaining **medically plausible**.  

- Dilute keywords and disperse critical information across the query.

- It must still be **answerable by the original answer**.  
\\[1em]
- Return **only the modified detailed query**.
    \end{tcolorbox}
    \caption{Claude 4 Opus prompt we used to ``distract'' questions. The detailed question is the full question from \autoref{fig:prompt_query} output. The concise question is the part that is necessary to derive the answer, while a detailed question usually conveys details unnecessary to reach the answer. Concise question remains semantically unchanged while other details are notably modified. }
    \label{fig:prompt_distractions}
\end{figure}
\begin{figure}[h!]
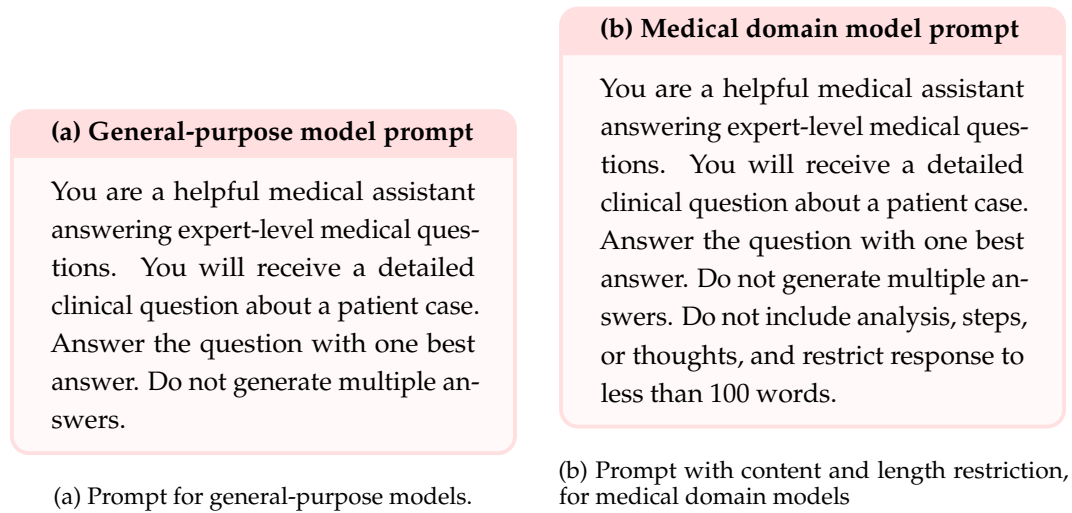

    \centering
    \begin{subfigure}{0.48\textwidth}
        \centering
        \begin{tcolorbox}[colframe=pink!50!white, colback=pink!10!white,
          boxrule=0.5mm, width=\textwidth, arc=2mm, auto outer arc,
          title=(a) General-purpose model prompt, fonttitle=\color{black}\bfseries]
          \setstretch{1.3}
          You are a helpful medical assistant answering expert-level medical questions. You will receive a detailed clinical question about a patient case. Answer the question with one best answer. Do not generate multiple answers.
        \end{tcolorbox}
        \caption{Prompt for general-purpose models.}
        \label{fig:prompt_a}
    \end{subfigure}
    \hfill
    \begin{subfigure}{0.48\textwidth}
        \centering
        \begin{tcolorbox}[colframe=pink!50!white, colback=pink!10!white,
          boxrule=0.5mm, width=\textwidth, arc=2mm, auto outer arc,
          title=(b) Medical domain model prompt, fonttitle=\color{black}\bfseries]
          \setstretch{1.3}
          You are a helpful medical assistant answering expert-level medical questions. You will receive a detailed clinical question about a patient case. Answer the question with one best answer. Do not generate multiple answers. Do not include analysis, steps, or thoughts, and restrict response to less than 100 words.
        \end{tcolorbox}
        \caption{Prompt with content and length restriction, for medical domain models}
        \label{fig:prompt_b}
    \end{subfigure}
    \caption{Prompts for generating responses with the questions from \benchmark{}.}
    \label{fig:prompt_responses}
\end{figure}

\begin{figure}[h!]
    \centering
    \begin{tcolorbox}[colframe=pink!50!white, colback=pink!10!white, boxrule=0.5mm, width=\textwidth, arc=2mm, auto outer arc, title=GPT-5.2 prompt for evaluating answer and model response equivalence, fonttitle=\color{black}\bfseries]
    \setstretch{1.3}
You are an expert clinical decision maker. 
You will be given two texts:

Gold: the gold standard next step in medical procedure for a specific patient and situation

Response: a model's proposed next step for the patient and situation
\\[1em]
Your task: decide whether Gold and Response have matching medical intent. 
\\[1em]
Criteria are following:
\\[1em]
Equivalence:

Return "Equivalence" if Response intends same clinical procedure as Gold, even if:

- worded differently

- different descriptions of the same surgical step or technique (e.g., “laparoscopic cholecystectomy” vs “surgical removal of gallbladder”)

- minor details differ as long as the core action is the same
\\[1em]
Mismatch: 

Return "Mismatch" if Response and Gold differs in any of these high-impact ways:

- different action type (e.g., monitoring vs. intervention, surgery vs. medication)

- significantly different order of action. A minor deviance is allowed

- different major additional action that changes the plan (e.g., Gold: “order CT”; Response: “start lobectomy)
\\[1em]
Follow these steps: 
\\[1em]
1. Compare two core medical actions: 

Extract the core medical action(s) from Gold. Express them as concise medical actions (e.g., “start chemotherapy,” “perform lobectomy,” “order CT scan”). Ignore details such as dose, frequency, or surgical technique/words unless they fundamentally change the type of action.

Extract the core medical action(s) from Response. Express them as concise medical actions (e.g., “start chemotherapy,” “perform lobectomy,” “order CT scan”). Ignore details such as dose, frequency, or surgical technique/words unless they fundamentally change the type of action.

2. Compare action, target, and clinical intent.

3. If unsure, default to Mismatch.

4. The output format should be one word: "Equivalence" or "Mismatch".
\\[1em]
Do not output any other texts.
\\[1em]
Few-shot examples

[Omitted for brevity]

\end{tcolorbox}
    \caption{GPT-5.2 prompt we used to evaluate the answer and LLM equivalency. The few-shot examples are drawn from early version of the dataset.}
    \label{fig:prompt_eval}
\end{figure}
\begin{figure}[h!]
    \centering
    \begin{tcolorbox}[colframe=blue!50!white, colback=blue!10!white, boxrule=0.5mm, width=\textwidth, arc=2mm, auto outer arc, title=Instruction for the annotators, fonttitle=\color{black}\bfseries]
    \setstretch{1.3}
There are 7 columns in the spreadsheet: 

- \textbf{Title}: title of the case report where the query and answer were derived from.

- \textbf{Classification}: the topic of the case report.

- \textbf{Link}: link to the case report. If you have any confusion or want to review the case report for validation, please use this link. If the query is straightforward, you don’t need to validate with the case report.

- \textbf{Query}: question presenting the case that is similar to the case report and asking the next step at a potentially confusing decision point, related to the significant point, or the reason why the case report was written. The query will not be a direct iteration of the case report, as we added distractions to make the query more interesting and difficult to answer.

- \textbf{Answer}: answer to the query, derived from the case report.

- \textbf{Rating}: your rating of the query-answer pair, in 1 to 5 scale.

- \textbf{Comments}: your comments about the query-answer pair. Feel free to leave it blank. 
\\[1em]
Rating and Comments columns are for you to fill with your opinions. We want to simulate a situation where a doctor is looking up a case report for reference, and whether the action will be taken or not is up to the doctor. Our goal is to test if the models are reliable for that purpose using our dataset. The gold answers on the spreadsheet reflect the answer presented by the case report, whether it is the standard or not.  We want to confirm that our information extraction was successful and that our queries are not too easy or obvious and correct. 
\\[1em]
To serve these purposes, the rating should be based on these criteria:
  
- The answer should be answering the query.

- The query should not be asking too “easy” question. The query should require knowledge from medical professionals to be answered.

- The provided answer is one of the most plausible acceptable answer given clinical context, ideally, the only answer. We understand that there will be multiple possible medical procedures given the characteristics of the medical field, but we want the answer provided to be at least one of the most reasonable answers. 

- There are some hallucinated data (e.g., answer is already stated on the question). We did our best to filter them using a model, but if you see malformed data, feel free to rate it as 1 without further consideration. 
\\[1em]
These are okay:

- The answer is not gold standard (The goal is to create a dataset reflecting the case report’s action taken.)

- The answer is not detailed 

- The answer is too specific to the patient (Some information, such as numbers, may be too specific to the patient, but the doctor potentially searching this case will also see the same information.)
    \end{tcolorbox}
    \caption{Instruction given to three annotators to verify question-answer pairs.}
    \label{fig:instructions}
\end{figure}

\clearpage
\section{Case Report Venues}\label{app:journal_names}
\begin{table}[h]
\small
\centering
\begin{tabular}{p{0.3\linewidth} p{0.3\linewidth} p{0.3\linewidth}}
\toprule
Case Reports Plast Surg Hand Surg & Case Rep Cardiol & SAGE Open Med Case Rep \\
Int J Med Pharm Case Reports      & JACC Case Rep    & Case Rep Genet \\
J Clin Cases Rep                  & Case Rep Clin Med & Case Rep Dent \\
J Clin Case Rep                   & Case Rep Surg    & Int Med Case Rep J \\
Case Rep Pancreat Cancer          & Trauma Case Rep  & Case Rep Ophthalmol \\
Arch Clin Case Rep                & Case Rep Pulmonol & Autops Case Rep \\
Respir Med Case Rep               & Clin Med Insights Case Rep & J Med Case Reports \\
Case Rep Neurol Med               & Clin Pract Cases Emerg Med & Gynecol Oncol Case Rep \\
Case Rep Transplant               & Case Rep Endocrinol & Clin Med Rev Case Rep \\
J Vasc Surg Cases Innov Tech      & Med Case Rep Short Rev & Psychiatry Res Case Rep \\
Clin Nephrol Case Stud            & Clin Case Rep Rev & Gen Thorac Cardiovasc Surg Cases \\
Arch Med Case Rep                 & Case Rep Emerg Med & MOJ Clin Med Case Rep \\
Endocrinol Diabetes Metab Case Rep & Prof Case Manag  & IDCases \\
J Cardiol Case Reports            & Ann Clin Case Rep & J Cardiol Cases \\
Am J Med Case Rep                 & Case Reports Immunol & Spinal Cord Ser Cases \\
Med Mycol Case Rep                & Int Clin Med Case Rep J & Oxf Med Case Reports \\
Case Rep Psychiatry               & IJU Case Rep     & Case Rep Otolaryngol \\
J Surg Tech Case Rep              & JCEM Case Rep    & Case Rep Ophthalmol Med \\
Clin Med Case Rep                 & Case Rep Dermatol & JAAD Case Rep \\
Ann Clin Med Case Rep             & J Pediatr Surg Case Rep & ACG Case Rep J \\
Surg Case Rep                     & Am J Ophthalmol Case Rep & Case Rep Crit Care \\
Case Rep Orthop                   & Case Rep Vet Med & Clin Case Stud \\
Case Rep Perinat Med              & GMS Ophthalmol Cases & CASE (Phila) \\
Case Rep Radiol                   & Case Rep Womens Health & Eur Heart J Case Rep \\
Open J Clin Med Case Rep          & Case Rep Gastrointest Med & Case Rep Infect Dis \\
Case Stud Eng Fail Anal           & Case Rep Oncol Med & Cases J \\
BJR Case Rep                      & J Surg Case Rep & Case Stud Chem Environ Eng \\
Case Rep Dermatol Med             & Clin Case Rep   & Indian J Ophthalmol Case Rep \\
Case Rep Pediatr                  & BMJ Case Rep    & Urol Case Rep \\
CEN Case Rep                      & Case Rep Anesthesiol & J Med Case Rep \\
Case Rep Urol                     & Case Reports Hepatol & Int J Surg Case Rep \\
Case Rep Obstet Gynecol           & Int J Case Rep Imag & J Investig Med High Impact Case Rep \\
Asploro J Biomed Clin Case Rep    & Case Rep Nephrol & Case Rep Hematol \\
Respirol Case Rep                 & Case Rep Pathol & Int J Clin Case Rep Rev \\
HeartRhythm Case Rep              & Neurocase       & AACE Clin Case Rep \\
Retin Cases Brief Rep             & Cold Spring Harb Mol Case Stud & Case Stud Transp Policy \\
JBJS Case Connect                 & J Endourol Case Rep & Case Rep Rheumatol \\
Case Rep Med                      & Oral Health Case Rep & Arch Clin Med Case Rep \\
JMM Case Rep                      & Radiol Case Rep & Epilepsy Behav Case Rep \\
Case Rep Vasc Med                 & APSP J Case Rep &  \\
\bottomrule
\end{tabular}
\caption{Journal names of the corpus of 53,617 case reports. Extracted from PMC commercially available file list provided by: \url{https://ftp.ncbi.nlm.nih.gov/pub/pmc/oa_non_comm_use_pdf.csv}}
\label{tab:venues}
\end{table}

\end{document}